%% file: main.tex
\documentclass[twocolumn]{svjour3}  
 
\usepackage[authoryear]{natbib}
\usepackage{graphicx}%
\usepackage{multirow}%
\usepackage{amsmath,amssymb,amsfonts}%
\usepackage{amsthm}%
\usepackage{mathrsfs}%
\usepackage[title]{appendix}%
\usepackage{xcolor}%
\usepackage{textcomp}%
\usepackage{manyfoot}%
\usepackage{booktabs}%
\usepackage{algorithm}%
\usepackage{algorithmicx}%
\usepackage{algpseudocode}%
\usepackage{listings}%

\usepackage{url}
\usepackage{booktabs}
\newcommand{\etal}{et al.}
\usepackage{xcolor}
\usepackage{multicol}

\usepackage{xcolor,colortbl}
\usepackage{makecell}
\usepackage{color}
\usepackage{tikz}
\usepackage[edges]{forest}
\definecolor{hidden-draw}{RGB}{20,68,106}
\definecolor{hidden-pink}{RGB}{255,245,247}
\usepackage{placeins}
\usepackage{rotating}
\usepackage{longtable}

\usepackage{tabularx}
\definecolor{tabblue}{HTML}{5555CC}

\usepackage{hyperref} 
\hypersetup{
hidelinks,
colorlinks=true,
linkcolor=tabblue,
urlcolor=tabblue,
citecolor=tabblue
}

\journalname{Journal of XXXX}

\begin{document}

\title{Diffusion Models in 3D Vision: A Survey}

\titlerunning{Diffusion Models in 3D Vision: A Survey}


\author{Zhen Wang \and 
        Dongyuan Li \and 
        Yaozu Wu \and 
        Tianyu He \and 
        Jiang Bian \and 
        Renhe Jiang$^{*}$}       

\institute{
Zhen Wang, Dongyuan Li, Yaozu Wu, Renhe Jiang \at 
Centre for Spatial Information Science, University of Tokyo, Tokyo, Japan \\ 
\email{zhenwangrs@gmail.com, lidy@csis.u-tokyo.ac.jp, yaozuwu279@gmail.com, jiangrh@csis.u-tokyo.ac.jp} 
\and
Tianyu He, Jiang Bian \at 
Microsoft Research Asia (MSRA), Beijing, China \\ 
\email{\{tianyuhe, jiang.bian\}@microsoft.com} \and
*Corresponding author.
}






\maketitle


\abstract{
In recent years, 3D vision has become a crucial field within computer vision, powering a wide range of applications such as autonomous driving, robotics, augmented reality, and medical imaging. 
This field relies on accurate perception, understanding, and reconstruction of 3D scenes from 2D images or text data sources. Diffusion models, originally designed for 2D generative tasks, offer the potential for more flexible, probabilistic methods that can better capture the variability and uncertainty present in real-world 3D data. 
In this paper, we review the state-of-the-art methods that use diffusion models for 3D visual tasks, including but not limited to 3D object generation, shape completion, point-cloud reconstruction, and scene construction. 
We provide an in-depth discussion of the underlying mathematical principles of diffusion models, outlining their forward and reverse processes, as well as the various architectural advancements that enable these models to work with 3D datasets. 
We also discuss the key challenges in applying diffusion models to 3D vision, such as handling occlusions and varying point densities, and the computational demands of high-dimensional data. 
Finally, we discuss potential solutions, including improving computational efficiency, enhancing multimodal fusion, and exploring the use of large-scale pretraining for better generalization across 3D tasks. This paper serves as a foundation for future exploration and development in this rapidly evolving field.
}




\input{Section/section1_Introduction}

\input{Section/section2_diffusion}

\input{Section/section3_3D}

\input{Section/section4_tasks}

\input{Section/section5_datasets}

\input{Section/section6_future}

\input{Section/section7_conclusion}


\bibliographystyle{bst/sn-basic}
\bibliography{main}


\end{document}

%% file: Section/section1_Introduction.tex
\section{Introduction}


Over the past few years, 3D vision \citep{3dvisionsurvey} has emerged as a vital area in computer vision, enabling a diverse array of applications, such as autonomous driving~\citep{zhang2024vision,kitti}, robotics~\citep{efficientdreamer,seolet}, augmented reality~\citep{diff_aug_depth,amparore2024computer}, and medical imaging~\citep{lai2024e3d,hatamizadeh2022unetr}. 
These applications rely on accurate perception, understanding, and reconstruction of 3D scenes from 2D data sources, such as images and videos. As 3D vision tasks become more complex, traditional methods often struggle with efficiency and scalability~\citep{DBLP:journals/pami/ZhengYZWCCL24,DBLP:journals/pami/HuangACZXN24}.

Diffusion models, originally proposed in the field of generative modeling, have rapidly evolved and show significant potential in many areas of computer vision~\citep{ddpm,score_based_gen,score_sde,improved_ddpm,learning_fast_sample}. 
These diffusion models, based on the idea of transforming data through a series of stochastic steps, have been successful in various tasks, such as image generation \citep{diffusion_beat_gan,high_img_syn_ldm,photorealistic,glide} and restoration tasks \citep{diffu_img_denoise,diffu_img_restore}. 
In particular, diffusion models have demonstrated strong generative capabilities that allow the generation of high-quality, diverse outputs while maintaining robustness against noise~\citep{DBLP:journals/csur/YangZSHXZZCY24,DBLP:journals/corr/abs-2407-00783}.

In recent years, the development of diffusion models has shifted from 2D to more challenging 3D tasks \citep{3d_paintbrush,diffusion_sdf_3d,dtu}, including 3D object generation \citep{point_diffusion,tiger,dit3d}, shape completion \citep{diffu_sdf,pdr}, and point cloud reconstruction \citep{scaling_diffusion}, marking a new phase in both diffusion modeling and 3D vision.




Diffusion models are increasingly being adopted for 3D vision tasks because of their unique ability to model intricate data distributions and iteratively refine outputs by learning noise-to-signal transformations. These capabilities make them particularly suitable for 3D applications like shape synthesis, point cloud completion, and depth estimation \citep{3dshape2vecset, ecodepth}, where real-world data often contain ambiguities, missing regions, or sensor noise. Unlike traditional deterministic approaches, which struggle to represent the inherent uncertainty and diversity of 3D environments, diffusion models operate probabilistically. This allows them to generate multiple plausible solutions and adapt to incomplete or sparse input, better mimicking the variability observed in physical scenes.

The success of diffusion models in 2D image generation, known for producing high-fidelity results, has further motivated their extension to 3D vision. By leveraging their iterative denoising process, these models excel in scenarios requiring robustness to occlusions or partial data, such as reconstructing full 3D shapes from limited views. 
This survey aims to consolidate the latest advancements in applying diffusion models to 3D vision, highlighting the advantages they bring, such as enhanced flexibility, robustness to noise, and more accurate representation of complex 3D structures.




Although diffusion models show immense potential for advancing 3D vision, their adoption faces significant technical and practical challenges. 
\textbf{First}, 3D datasets are inherently limited in scale compared to 2D image repositories, which benefit from vast and easily accessible collections (e.g., photos or videos). This scarcity complicates training robust 3D diffusion models, as 3D data curation is labor-intensive and often domain-specific (e.g., medical scans or LiDAR point clouds).
\textbf{Second}, 3D data's structural complexity introduces unique hurdles: representations like meshes, voxels, or point clouds each require specialized architectural adaptations to integrate with diffusion frameworks. For example, meshes need noise modeling that considers their structure, while point clouds require processing methods that are unaffected by point order. These challenges are uncommon in 2D scenarios.
\textbf{Finally}, computational demands escalate dramatically in 3D tasks. Operations such as volumetric rendering or point-cloud diffusion involve high-dimensional computations, and the iterative nature of diffusion processes compounds these costs. Scaling such models to handle real-world 3D scenes while maintaining fidelity and efficiency remains an open problem.


This survey focuses on the application of diffusion models in a wide range of 3D vision tasks, including but not limited to 3D object generation \citep{renderdiffusion}, novel view synthesis \citep{generative_novel_view}, 3D editing \citep{datid3d}, and scene generation \citep{scene_lidar_diffu}. 
We review various diffusion model architectures and their adaptations to 3D vision, covering both early-stage and recent advancements made in the past five years. Special attention is paid to discuss how these models address the challenges specific to 3D data and the computational constraints associated with large-scale 3D vision problems.
The key contributions of this paper could be summarized as follows: 
\begin{itemize} 
    \item A comprehensive categorization and summary of existing work that uses diffusion models to 3D vision tasks, including their advantages and limitations. 
    \item An in-depth analysis and comparison of various key techniques, frameworks, and methodologies used to adapt diffusion models for 3D data. 
    \item A detailed discussion on the current challenges and open problems in the field, along with potential future research directions for improving diffusion models in 3D vision applications. 
    \item An extensive review of relevant datasets and benchmarks commonly used to evaluate the performance of diffusion models in various 3D vision tasks. 
\end{itemize}



For our survey, we employ a comprehensive literature search strategy to ensure a thorough exploration of the field. We begin by identifying key terms and phrases relevant to our topic, including ``diffusion models'', ``3D vision'', and related concepts such as ``generative models'' and ``neural networks for 3D data''. 
We conduct our search across multiple academic databases, including but not limited to IEEE Xplore \footnote{\url{https://ieeexplore.ieee.org/Xplore/home.jsp}}, arXiv \footnote{\url{https://arxiv.org/}}, and Google Scholar \footnote{\url{https://scholar.google.com/}}, focusing on publications from the last five years to capture recent advances. 
In addition, we prioritize peer-reviewed journal articles, conference papers, and preprints, ensuring the inclusion of high-quality and cutting-edge research. Using this strategy, our aim is to provide a comprehensive and up-to-date survey of diffusion models in 3D vision.

\begin{figure*}[h]
  \centering
  \includegraphics[width=0.99\textwidth]{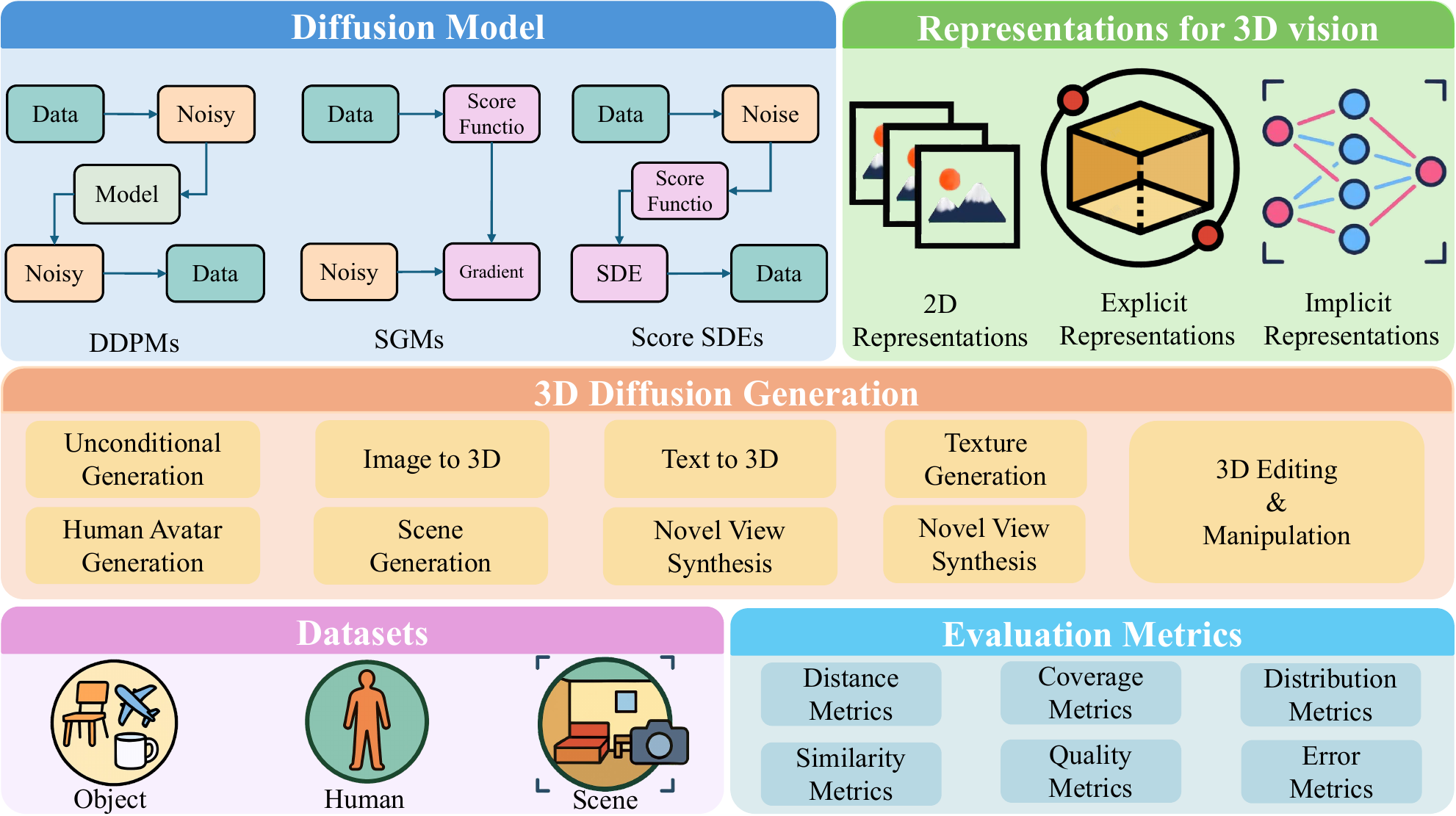}
  \caption{The overall framework of this survey.}
  \label{fig:intro}
\end{figure*}

As shown in Figure~\ref{fig:intro}, the remainder of the paper is organized as follows. Section \ref{sec:diffusion} provides an overview of diffusion models, including their theoretical foundations and key developments in 2D and 3D vision tasks. Section \ref{sec:3d} dives into the core concepts of 3D vision, discussing various data representations and their challenges. Section \ref{sec:tasks} presents a detailed review of diffusion models applied to different 3D vision tasks. In Section \ref{sec:dataset}, we summarize the available datasets and benchmarks used for the evaluation. Finally, Section \ref{sec:future} discusses future directions and open problems, followed by a conclusion in Section \ref{sec:conclusion}.

%% file: Section/section2_diffusion.tex
\section{Diffusion Model Basics}
\label{sec:diffusion}

In this Section, we first give an overview of diffusion models and then introduce various core diffusion models and how they can be applied to 3D Vision tasks.

\begin{figure}[h]
  \centering
  \includegraphics[width=\linewidth]{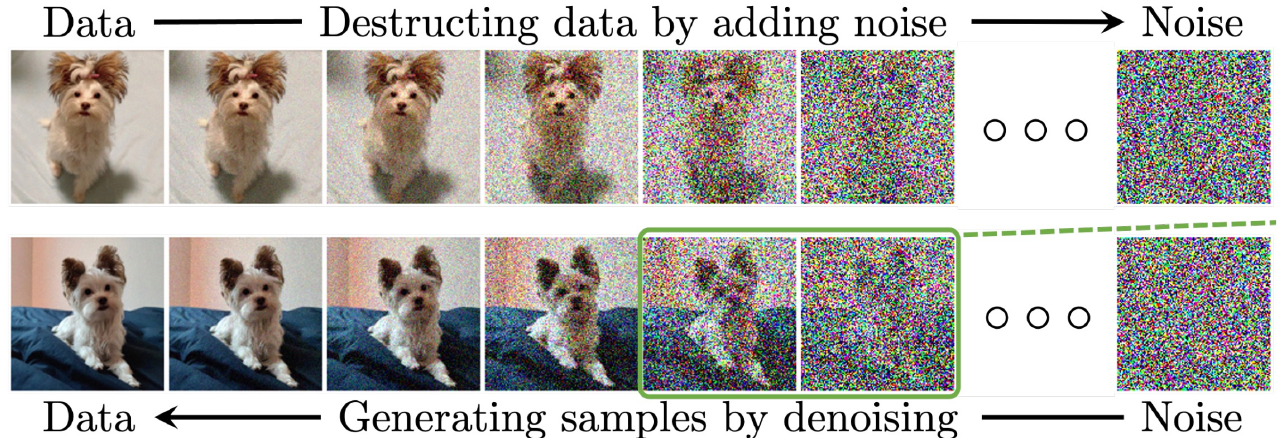}
  \caption{Diffusion models smoothly perturb data by adding noise, then reverse this process to generate new data from noise. Each denoising step in the reverse process typically requires estimating the score function~\citep{yang2023diffusion}.}
  \vspace{-30pt}
  \label{fig:illustration}
\end{figure}

\subsection{Diffusion Foundations}

\noindent \textbf{Definition.} As shown in Figure~\ref{fig:illustration}, diffusion models are a class of generative models that learn to generate data by gradually transforming random noise into structured data~\citep{ddpm}. 
This process involves a forward diffusion process, where noise is added step by step to the data, and a reverse process that denoises and reconstructs data from pure noise. 
Their purpose is to model the data distribution through iterative denoising to avoid mode collapse, a common issue in other generative models, such as GANs, and produce diverse high-quality samples. 
We then give a detailed introduction to the general framework of diffusion models.



\subsubsection{Forward Diffusion Process}

Forward diffusion process systematically perturbs the data distribution $q(\mathbf{x})$ through a discrete-time Markov chain with $T$ transitions.  
Given a clean data sample $\mathbf{x}_0 \sim q(\mathbf{x}_0)$, the noising process could be denoted as:
\begin{equation}
    q(\mathbf{x}_{1:T}|\mathbf{x}_0) = \prod_{t=1}^T q(\mathbf{x}_t|\mathbf{x}_{t-1}).
\end{equation}

Each noising process is defined by a transition kernel parameterized by $\{\beta_t\}_{t=1}^T$. The $t$-th step of the forward diffusion process could then be denoted as follows:
\begin{equation}
    q(\mathbf{x}_t|\mathbf{x}_{t-1}) = \mathcal{N}\left(\mathbf{x}_t; \sqrt{1-\beta_t}\mathbf{x}_{t-1}, \beta_t \mathbf{I}\right),
\end{equation}
where $\beta_{t} \in (0,1)$ is a pre-defined noise schedule controlling the noise intensity at step $t$, and $\mathbf{I}$ denotes the identity matrix. Using the reparameterization trick and the product rule for Gaussian distributions~\citep{kingma2015variational}, the cumulative transition probability from $\mathbf{x}_{0}$ to $\mathbf{x}_{t}$ can be defined as follows:
\begin{equation}
    q(\mathbf{x}_t|\mathbf{x}_0) = \mathcal{N}\left(\mathbf{x}_t; \sqrt{\alpha_t} \mathbf{x}_0, (1 - \alpha_t) \mathbf{I}\right),
\end{equation}
where $\alpha_t := \prod_{s=1}^t (1-\beta_s)$ satisfies $\lim_{t\to T} \alpha_t \approx 0$, ensuring the distribution $q(\mathbf{x}_T)$ approaches $\mathcal{N}(\mathbf{0}, \mathbf{I})$, effectively removing information from the original data while preserving analytical tractability.

\subsubsection{Reverse Diffusion Process}
Given that the forward diffusion process introduces noise in a controlled manner, the reverse diffusion process aims to progressively remove this noise and reconstruct a clean data sample $\mathbf{x}_{0}$ starting from a Gaussian prior. This reconstruction is achieved by learning a denoising transition function $p_\theta(\mathbf{x}_{t-1}|\mathbf{x}_t)$ that approximates the true posterior distribution as follows:
\begin{equation}
    p_\theta(\mathbf{x}_{0:T}) = p(\mathbf{x}_T) \prod_{t=1}^T p_\theta(\mathbf{x}_{t-1}|\mathbf{x}_t),
\end{equation}
where the initial distribution is defined as $p(\mathbf{x}_T) = \mathcal{N}(\mathbf{0}, \mathbf{I})$. At each reverse diffusion step, the learned transition distribution $p_\theta(\mathbf{x}_{t-1}|\mathbf{x}_t)$ approximates the true posterior distribution $q(\mathbf{x}_{t-1}|\mathbf{x}_t, \mathbf{x}_0)$ as follows:
\begin{equation}
    p_\theta(\mathbf{x}_{t-1}|\mathbf{x}_t) = \mathcal{N}\left(\mathbf{x}_{t-1}; \boldsymbol{\mu}_\theta(\mathbf{x}_t, t), \sigma_t^2 \mathbf{I}\right).
\end{equation}
where $\boldsymbol{\mu}_\theta(\mathbf{x}_t, t)$ and $\sigma_t^2 \mathbf{I}$ denote the learned mean and covariance of the reverse transitions, respectively. The model progressively denoises the perturbed samples via iterative refinement, ultimately approximating the original data distribution $p_{\theta}(\mathbf{x}_{0})$. 

In practical implementations, the variance $\sigma_t^2$ is often fixed or defined as a deterministic function of $\beta_{t}$ to improve the stability of training. Furthermore, neural networks are commonly used to parameterize the mean $\boldsymbol{\mu}_\theta(\mathbf{x}_t, t)$, typically directly predicting the added noise term $\boldsymbol{\epsilon}$.
For example, \cite{ddpm} set $\sigma_t^2 = \beta_t$ and parameterize the mean via a neural network $\boldsymbol{\epsilon}_\theta$ that explicitly predicts the noise:
\begin{equation}
    \boldsymbol{\mu}_\theta(\mathbf{x}_t, t) = \frac{1}{\sqrt{1-\beta_t}}\left(\mathbf{x}_t - \frac{\beta_t}{\sqrt{1-\alpha_t}} \boldsymbol{\epsilon}_\theta(\mathbf{x}_t, t)\right),
\end{equation}
where this reparameterization decouples the task of data reconstruction from noise estimation, leading to improved training stability than predicting the mean.

\subsubsection{Probability Density and Score Matching}

Diffusion models typically approximate the data distribution through score-based generative modeling, a framework that learns the gradient of the log probability density, known as the \textit{score function}~\citep{robbins1992empirical}.
Formally, for a perturbed sample $\mathbf{x}_t \sim q(\mathbf{x}_t|\mathbf{x}_0)$, the score function could be defined as: 
\begin{equation} 
\footnotesize
\nabla_{\mathbf{x}_{t}} \log p_{\theta}(\mathbf{x}_{t}) = \mathbb{E}_{q(\mathbf{x}_t | \mathbf{x}_0)}\left[\frac{\sqrt{\alpha_{t}}\mathbf{x}_{0}-\mathbf{x}_{t}}{1-\alpha_{t}}\right] = - \frac{\boldsymbol{\epsilon}_{\theta} (\mathbf{x}_{t}, t)}{\sqrt{1-\alpha_{t}}},
\end{equation}    
where $\boldsymbol{\epsilon}_{\theta}$ represents a scaled estimator of the score function. As the true distribution of data $q(\mathbf{x}_{t})$ is typically unknown, the reverse diffusion process approximates the score function with a parameterized model. Specifically, to ensure consistency between the true and estimated score functions, the score matching (SM) loss could be formulated as follows:
\begin{footnotesize}
\begin{equation}\label{eq.8}
\mathcal{L}_{\text{SM}} = \mathbb{E}_{q(\mathbf{x}_t | \mathbf{x}_0)} \left[ \left\| \nabla_{\mathbf{x}_t} \log q(\mathbf{x}_t | \mathbf{x}_0) - \nabla_{\mathbf{x}_t} \log p_\theta(\mathbf{x}_t) \right\|^2_{2} \right].
\end{equation}
\end{footnotesize}

This objective is closely related to variational inference, as minimizing it corresponds to reducing the Fisher divergence between the true diffusion process and the learned model distribution, thus ensuring an accurate reconstruction of the original data.
By substituting the relationship of the score function and utilizing the law of total expectation (tower property), Eq.~\eqref{eq.8} can be reformulated as follows:
\begin{equation}
    \mathcal{L}_{\text{SM}} = \mathbb{E}_{q(\mathbf{x}_t|\mathbf{x}_0)}\left[ \frac{\lambda(t)}{2} \left\| \boldsymbol{\epsilon}_\theta - \frac{\mathbf{x}_t - \sqrt{\alpha_t}\mathbf{x}_0}{\sqrt{1-\alpha_t}} \right\|^2_{2} \right],
\end{equation}
where $\lambda(t) = (1-\alpha_t)/\alpha_t$ compensates for the varying scale of variance across different diffusion steps.

\subsubsection{Overall Loss Function}
Building upon the previously defined score matching objective, the primary training goal of diffusion models is minimizing \textit{denoising score matching loss}, which encourages accurate prediction of the added noise:
\begin{equation}
\mathcal{L}_{\text{Denoise}} = \mathbb{E}_{t, \mathbf{x}_0, \boldsymbol{\epsilon}} \left[ \left\| \boldsymbol{\epsilon} - \boldsymbol{\epsilon}_\theta(\mathbf{x}_t, t) \right\|^2 \right],
\end{equation}
where $\boldsymbol{\epsilon}_\theta(\mathbf{x}_t, t)$ represents the model’s prediction of the noise $\boldsymbol{\epsilon}$ added at timestep $t$, and $t$ is uniformly sampled from the diffusion timesteps.

Importantly, this loss function directly aligns with the previously introduced score matching objective, since predicting noise implicitly captures the underlying score function of the data distribution. This approach offers notable theoretical and practical benefits: it simplifies optimization by circumventing explicit density estimation and naturally connects diffusion models with maximum likelihood estimation. Specifically, minimizing $\mathcal{L}_{\text{Denoise}}$ corresponds to maximizing a variational lower bound of the log-likelihood of the data distribution, thus providing a rigorous probabilistic foundation for training. Furthermore, uniform sampling of the timestep $t$ ensures balanced training across all noise levels, improving the stability and robustness of the learned model throughout the diffusion process.

\subsection{Diffusion Taxonomy}

Current research on diffusion models is mainly based on three predominant formulations: denoising diffusion probabilistic models (DDPMs)~\citep{ddpm,improved_ddpm,ddpm_nonequilibrium}, score-based generative models (SGMs)~\citep{score_sde}, and stochastic differential equations (Score SDEs)~\citep{song2021maximum}. 
We give a self-contained introduction to these three formulations while discussing their connections with each other along the way.

\begin{figure}[h]
  \centering
  \includegraphics[width=\linewidth]{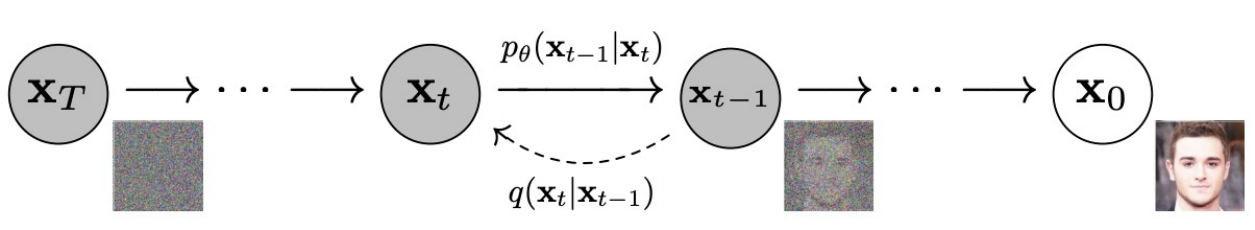}
  \caption{The directed graphical model considered in denoising diffusion probabilistic models~\citep{ddpm}.}
  \label{fig:ddpm}
  \vspace{-20pt}
\end{figure}

\subsubsection{Denoising Diffusion Probabilistic Models}

DDPMs represent one of the most prominent and widely studied classes of diffusion models~\citep{ddpm}. As illustrated in Figure~\ref{fig:ddpm}, the forward diffusion process systematically perturbs the data by incrementally adding Gaussian noise, eventually transforming the original data distribution into a tractable prior, typically an isotropic Gaussian distribution. Conversely, the reverse diffusion process, parameterized by a neural network, progressively removes this noise via iterative denoising steps, ultimately reconstructing the original data distribution.

Formally, given an initial data point drawn from the distribution $\mathbf{x}_0 \sim q(\mathbf{x}_0)$, the forward diffusion process defines a Markov chain that sequentially generates a set of latent variables $\mathbf{x}_1, \mathbf{x}_2, \dots, \mathbf{x}_T$, governed by the transition kernel $q(\mathbf{x}_t|\mathbf{x}_{t-1})$. Utilizing the Markov property and the chain rule of probability, the joint distribution of the latent variables conditioned on $\mathbf{x}_0$, denoted by $q(\mathbf{x}_{1:T}|\mathbf{x}_0)$, can be factorized as: 
    $q(\mathbf{x}_{1:T}|\mathbf{x}_0) = \prod_{t=1}^T q(\mathbf{x}_t|\mathbf{x}_{t-1}).$
In DDPMs, current studies typically specify the transition kernel $q(\mathbf{x}_t|\mathbf{x}_{t-1})$ to incrementally transform the data distribution $q(\mathbf{x}_0)$ into a tractable prior distribution. A common and effective choice for this kernel is Gaussian perturbation, formulated as follows:
    $q(\mathbf{x}_t | \mathbf{x}_{t-1}) = \mathcal{N}(\mathbf{x}_t; \sqrt{1 - \beta_t} \mathbf{x}_{t-1}, \beta_t \mathbf{I}), $
where $\mathbf{x}_t$ denotes the perturbed data at $t$-th step, and $\beta_t$ is a predefined variance schedule controlling the intensity of the injected noise in each step. As $t$ increases, the cumulative addition of noise ensures that $q(\mathbf{x}_{T})$ converges to an isotropic Gaussian distribution. As noted in \cite{sohl2015deep}, the Gaussian transition kernel allows the joint distribution to be analytically marginalized, providing a closed-form expression for $q(\mathbf{x}_t | \mathbf{x}_0)$ for all steps $t \in \{1,\dots,T\}$. Defining $\alpha_t := 1 - \beta_t$ and $\bar{\alpha}t := \prod{s=0}^{t} \alpha_s$, we obtain:
$q(\mathbf{x}_t|\mathbf{x}_0) = \mathcal{N}(\mathbf{x}_t; \sqrt{\bar{\alpha}_t} \mathbf{x}_0, (1-\bar{\alpha}_t) \mathbf{I}).$
Given $\mathbf x_0$, we can easily obtain a sample of $\mathbf{x}_t$ by sampling a Gaussian vector $\boldsymbol{\epsilon} \sim \mathcal{N}(\mathbf{0}, \mathbf{I})$ and applying the following transformation:
$\mathbf{x}_t =\sqrt{\bar{\alpha}_t} \mathbf{x}_0 + \sqrt{1-\bar{\alpha}_t} \boldsymbol{\epsilon}.$ 
where \(\bar{\alpha}_T \approx 0 \). Under this condition, the distribution of $\mathbf{x}_{T}$ becomes approximately Gaussian. Then, we have: 
$q(\mathbf x_T) := \int q(\mathbf x_T | \mathbf x_0) q(\mathbf x_0) \textrm{d} \mathbf x_0 \approx \mathcal{N}(\mathbf x_T; \mathbf{0}, \mathbf{I}),$
indicating that the forward diffusion process transforms the original data distribution into a Gaussian.

The forward process gradually injects noise into the data, progressively removing structural information until all meaningful structures are lost. To generate new data samples, DDPMs begin by sampling an unstructured noise vector from a predefined prior distribution and then progressively remove noise through a learned Markov chain that operates in the reverse direction.
Formally, this reverse Markov chain is defined by a prior distribution $p(\mathbf x_T) = \mathcal{N}(\mathbf x_T; \mathbf{0}, \mathbf{I})$ and a parameterized transition kernel $p_\theta(\mathbf{x}_{t-1}|\mathbf{x}_t)$. The choice of a prior distribution $p(\mathbf x_T) = \mathcal{N}(\mathbf x_T; \mathbf 0, \mathbf I)$ aligns with the property of the forward diffusion process, where $q(\mathbf x_T) \approx \mathcal{N}(\mathbf x_T; \mathbf{0}, \mathbf{I})$. The learned transition kernel is defined as:
$p_\theta(\mathbf x_{t-1}|\mathbf x_t) = \mathcal{N}(\mathbf x_{t-1}; \boldsymbol{\mu}_{\theta}(\mathbf x_t, t), \boldsymbol{\Sigma}_{\theta}(\mathbf x_t, t)), $
where $\theta$ denotes the parameters of deep neural networks that predict both the mean $\boldsymbol{\mu}_{\theta}(\mathbf x_t, t)$ and the covariance $\boldsymbol{\Sigma}_{\theta}(\mathbf x_t, t)$. With this reverse Markov chain, generating a data sample $\mathbf x_0$ involves first drawing a noise vector $\mathbf x_T \sim p(\mathbf x_T)$, and then iteratively sampling through the learned transition kernel as: 
$\mathbf x_{t-1} \sim p_\theta(\mathbf x_{t-1} | \mathbf x_t), t \in \{T, T-1, \ldots, 1\}$

\cite{ho2020denoising} propose reweighting the terms in the variational lower bound (VLB) to enhance sample quality. They highlight an important equivalence between their reweighted objective and the training loss used by noise-conditional score networks, a variant of \textit{score-based generative models}~\citep{song2019generative}. The resulting loss function can be formulated as follows: 
$\mathbb{E}_{t,\mathbf{x}_0,\boldsymbol{\epsilon}}\left[{ \lambda(t)  \left\| \boldsymbol{\epsilon} - \boldsymbol{\epsilon}_\theta(\mathbf{x}_t, t) \right\|^2_{2}}\right],$
where $\lambda(t)$ is a positive weighting function, the noisy sample $\mathbf x_t$ is calculated from the clean data $\mathbf x_0$ and the noise \(\boldsymbol{\epsilon}_{\theta}\) is a neural network parameterized by $\theta$, designed to predict the noise vector \(\boldsymbol{\epsilon}\) from inputs \(\mathbf{x}_{t}\) and $t$. 

\subsubsection{Score-Based Generative Models}

Score-based generative models (SGMs)~\citep{score_based_gen,score_sde} rely on explicitly estimating \emph{score function}, which quantifies how rapidly a probability density function changes at each data point. Formally, for a given probability density function $p(\mathbf{x})$, the score function is defined as the gradient of its log-density:
$\nabla_{\mathbf{x}} \log p(\mathbf{x})$. 
Unlike the Fisher score $\nabla_{\theta}\log p_{\theta}(\mathbf{x})$, frequently used in statistical inference and defined with respect to the model parameters $\theta$, the score function in SGMs is directly associated with the data points $\mathbf{x}$. It identifies the directions along which the probability density increases most rapidly.

The fundamental principle behind score-based generative models is to progressively perturb the data by adding Gaussian noise of increasing intensity. Formally, given an initial data distribution $q(\mathbf{x}_0)$ and a series of increasing noise levels $0 < \sigma_1 < \sigma_2 < \dots < \sigma_T$, each data point $\mathbf{x}_0$  is transformed into noisy samples $\mathbf{x}_t$, according to:
$q(\mathbf{x}_t|\mathbf{x}_0)=\mathcal{N}(\mathbf{x}_t;\mathbf{x}_0,\sigma_t^2\mathbf{I}),$
leading to a sequence of noisy distributions defined by:
$q(\mathbf{x}_t)=\int q(\mathbf{x}_t|\mathbf{x}_0)q(\mathbf{x}_0)d\mathbf{x}_0. $
Subsequently, a deep neural network called the {noise-conditional score network}, denoted as $\mathbf{s}_\theta(\mathbf{x}_t,t)$, is trained to approximate the score function $\nabla_{\mathbf{x}_t}\log q(\mathbf{x}_t)$ on each noise scale. A widely adopted training strategy is {denoising score matching}, which minimizes 
$\mathbb{E}_{t,\mathbf{x}_0,\boldsymbol{\epsilon}}\left[\lambda(t)\|\boldsymbol{\epsilon}+\sigma_t\mathbf{s}_\theta(\mathbf{x}_t,t)\|^2\right],$
where $\mathbf{x}_t=\mathbf{x}_0+\sigma_t\boldsymbol{\epsilon}$, $\boldsymbol{\epsilon}\sim\mathcal{N}(\mathbf{0},\mathbf{I})$, and $\lambda(t)$ is a positive weighting factor. 
Notably, this objective closely aligns SGMs with denoising diffusion probabilistic models. Setting $\boldsymbol{\epsilon}_\theta(\mathbf{x}_t,t)=-\sigma_t\mathbf{s}_\theta(\mathbf{x}_t,t)$ reveals an equivalence between their training objectives. Recent advances~\citep{meng2021estimating} further extend score matching by incorporating higher-order derivatives to improve sampling efficiency. 

After training, SGMs generate new data samples by iteratively refining initial noisy inputs using the learned score functions across decreasing noise levels. 
A prominent sampling method is annealed Langevin dynamics, which initializes with pure Gaussian noise  $\mathbf{x}_T^{(N)}\sim\mathcal{N}(\mathbf{0},\mathbf{I})$, and iteratively refines the samples via: 
$\boldsymbol{\epsilon}^{(i)} \sim\mathcal{N}(\mathbf{0},\mathbf{I}),
\mathbf{x}_t^{(i+1)} \leftarrow\mathbf{x}_t^{(i)}+\frac{s_t}{2}\mathbf{s}_\theta(\mathbf{x}_t^{(i)},t)+\sqrt{s_t}\boldsymbol{\epsilon}^{(i)}, $
where $s_t>0$ denotes the step size, $N$ is the number of iterations per noise level and $t=\{T,\dots,1\}$. 

\begin{figure}[h]
  \centering
  \includegraphics[width=\linewidth]{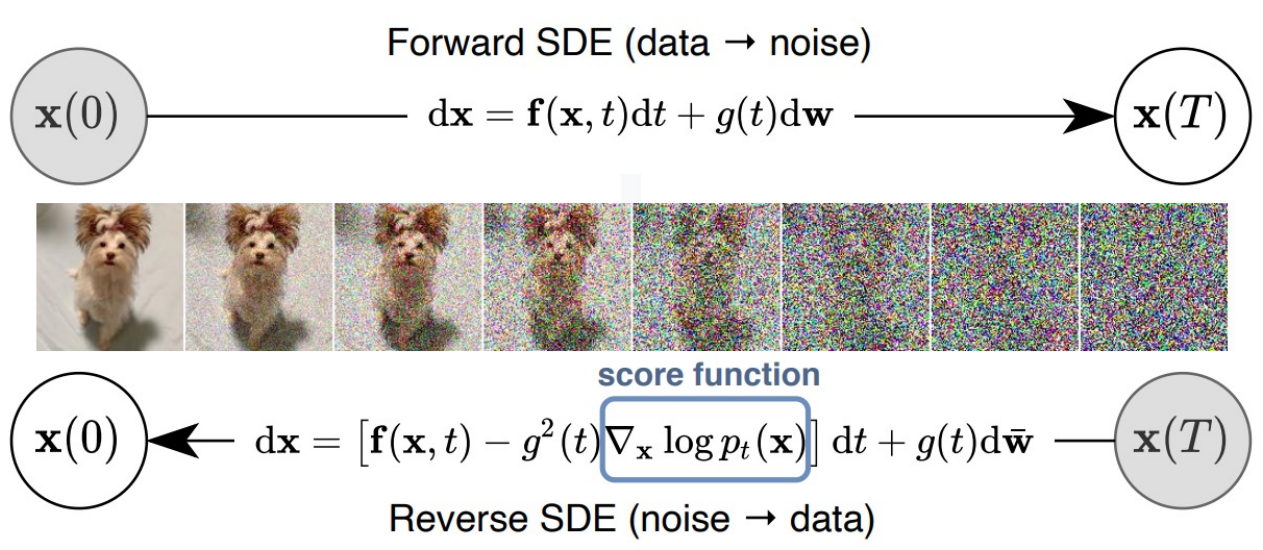}
  \caption{Solving a reversetime stochastic differential equations yields a score-based generative model. Transforming data to a simple noise distribution can be accomplished with a continuous-time SDE~\citep{score_sde}.}
  \label{fig:score}
  \vspace{-10pt}
\end{figure}

\subsubsection{Stochastic Differential Equations}

Stochastic Differential Equations~\citep{stochasti_differential} offer a continuous-time formulation of diffusion models, providing a more flexible alternative compared to the discrete-time approach used in Denoising Diffusion Probabilistic Models. Rather than evolving data through a fixed sequence of discrete timesteps, SDE-based methods continuously transform data distributions over time, enabling adaptive and fine-grained control of the diffusion process.

As shown in Figure~\ref{fig:score}, the forward process in this framework is governed by an Itô SDE, which models the evolution of data and can be formulated as follows:
    $d\mathbf{x} = f(\mathbf{x}, t) dt + g(t) d\mathbf{w},$
where $f(\mathbf{x}, t)$ is the drift function that dictates deterministic motion, $g(t)$ is the diffusion coefficient controlling stochastic variation, and $d\mathbf{w}$ denotes the Wiener process (Brownian motion) responsible for injecting noise over continuous time. This formulation generalizes discrete-time diffusion models by smoothly degrading the data distribution. Specifically, as shown by \cite{song2020score}, the corresponding SDE for DDPMs is:
    $\textrm{d} \mathbf{x} = -\frac{1}{2}\mathbf{\beta}(t)\mathbf{x}dt + \sqrt{\beta(t)}\textrm{d}\mathbf{w} $
where \(\mathbf{\beta}(\frac{t}{T}) = T\beta_{t}\) as $T$ goes to infinity. Similarly, for score-based generative models, the corresponding SDE is:
    $\textrm{d} \mathbf{x} = \sqrt{\frac{\textrm{d} [\sigma(t)^2]}{\textrm{d} t}}\textrm{d}\mathbf{w} \label{eqn:sgmsde}, $
where $\sigma(\frac{t}{T}) = \sigma_t$ as $T$ goes to infinity. We denote the resulting marginal distribution of $\mathbf{x}_t$ at step $t$ by $q_t(\mathbf{x})$. Crucially, Anderson's result~\citep{anderson1982reverse} ensures that for any forward diffusion process, the reverse-time dynamics is governed by the reverse-time SDE: 
$    d\mathbf{x} = [f(\mathbf{x}, t) - g(t)^2 \nabla_{\mathbf{x}} \log p_t(\mathbf{x})] dt + g(t) d\bar{\mathbf{w}}, $
where $\bar{\mathbf{w}}$ is a standard Wiener process when time flows backward and $\textrm{d}t$ denotes an infinitesimal negative time step. The trajectory of this reverse SDE shares marginal densities identical to those of the forward SDE. Intuitively, this reverse-time formulation describes a diffusion process that gradually transforms noise into meaningful data.

Moreover, \cite{song2020score} prove the existence of an ordinary differential equation, namely the probability flow ODE, whose trajectories have the same marginals as the reverse-time SDE. The probability flow ODE is given by:
  $  \textrm{d} \mathbf{x} = \left[{f}(\mathbf{x},t) - \frac{1}{2} g(t)^2\nabla_{\mathbf{x}} \log q_t(\mathbf{x})\right] \textrm{d}t.$
Both the reverse-time SDE and the probability flow ODE enable sampling from the true data distribution, as their solution trajectories share the same marginal densities. Given the score function $\nabla_{\mathbf{x}} \log q_t(\mathbf{x})$ at each timestep $t$, we can solve these equations numerically to generate new data samples. Practically, similar to standard SGMs, we parameterize a time-dependent score network \(\mathbf{s}_{\theta}(\mathbf{x}_t,t)\) to approximate this score function. Generalizing the discrete-time score matching objective to continuous time yields the following training objective: 
    $\mathbb{E}_{t,\mathbf{x}_0,\boldsymbol{\epsilon}} \left[ \lambda(t)  \left\| \mathbf{s}_{\theta}(\mathbf{x}_t,t) - \nabla_{\mathbf{x}_t} \log q(\mathbf{x}_t | \mathbf{x}_0) \right\|^2\right]. $

\subsubsection{Other Variants}



Beyond standard diffusion models, some extensions enhance flexibility and applicability. Hybrid methods combine discrete-time DDPMs and continuous-time SDEs, leveraging their respective advantages. Other models replace the Gaussian noise assumption with mixtures or alternative stochastic processes to improve generative performance on diverse data, including text and structured data. Conditional diffusion models~\citep{conditional_ddpm} introduce auxiliary inputs like class labels or semantic attributes to guide synthesis. These models excel in controlled generation tasks, notably conditional 3D object generation, scene completion, and cross-modal synthesis, showcasing their versatility for complex structured data.

%% file: Section/section3_3d.tex
\section{3D Vision Fundamentals}
\label{sec:3d}

\subsection{Representations for 3D Vision}

\begin{figure*}[h]
  \centering
  \includegraphics[width=\linewidth]{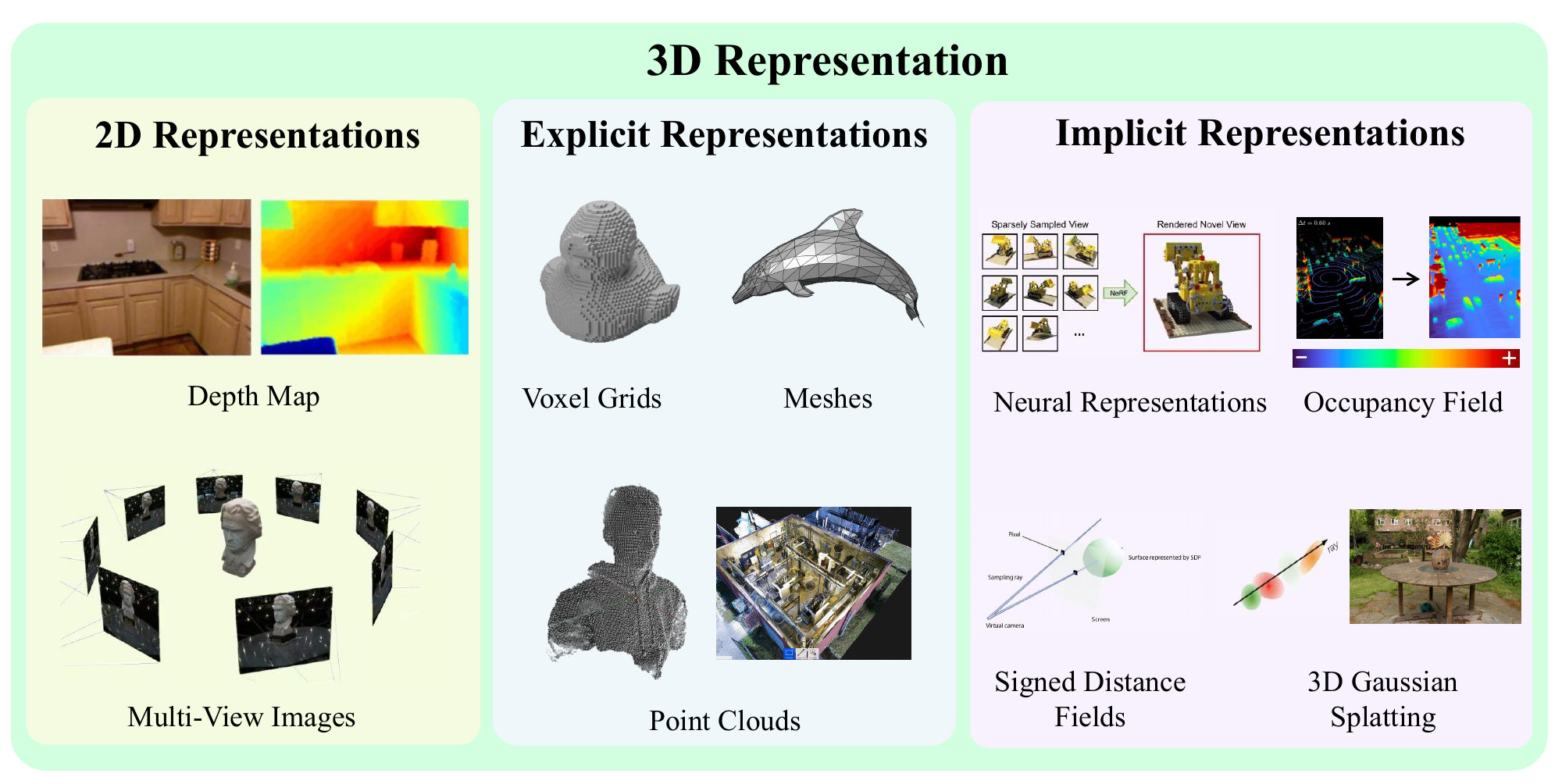}
  \caption{3D representations.}
  \label{fig:3d_data}
\end{figure*}

Three-dimensional data representation forms the cornerstone of 3D vision, providing the means to model, analyze, and interact with complex spatial information. In an era where applications such as augmented reality (AR), virtual reality (VR), and advanced robotics are increasingly prevalent, a robust understanding of 3D representation is essential for interpreting and manipulating the physical world. The representation of 3D data can take various forms, each with distinct characteristics, advantages, and limitations.

\subsubsection{2D Representations for 3D Vision}

2D representations use two-dimensional images to infer and reconstruct three-dimensional information. 
These representations are particularly useful for rendering and understanding 3D scenes but rely heavily on the quality and number of 2D inputs for accuracy.

\paragraph{Depth Map}

A depth map \citep{nyudepth} represents the distance from a particular viewpoint (often a camera) to objects in the scene as a 2D image, where each pixel's value corresponds to the depth (or distance) to the nearest surface. Depth maps are commonly generated from stereo vision, structured light, or time-of-flight (ToF) cameras, which estimate depth based on the geometry of light or motion.
Depth maps offer an efficient way to encode 3D information in a format that is easy to integrate with traditional 2D image processing techniques. They are particularly useful in applications like 3D reconstruction, scene understanding, and robotic navigation, where understanding the spatial layout of a scene is essential. In addition, depth maps can be used in conjunction with RGB images to create RGB-D datasets, which combine both color and depth information for more comprehensive scene analysis.
One limitation of depth maps is that they typically represent only the visible surfaces of objects from a single viewpoint. This means that occluded areas (i.e., parts of objects that are hidden from the camera) are not captured, which can pose challenges for certain 3D reconstruction tasks.

\paragraph{Multi-View Images}

Multi-view images \citep{dtu,llff} refer to multiple 2D images of the same scene or object taken from different perspectives. These images can be used to infer 3D information using techniques like stereo matching or structure-from-motion (SfM). The idea is to leverage the differences in appearance across views (known as parallax) to recover the geometry of the scene.
Multi-view images are widely used in 3D reconstruction, virtual reality (VR), and augmented reality (AR) applications, where they help create immersive, real-world-like experiences. In photogrammetry, for instance, multiple overlapping images of a scene are analyzed to generate detailed 3D models. Similarly, in autonomous driving, multi-view images from multiple cameras mounted on a vehicle are used to perceive the surrounding environment in 3D.
Compared to other 3D representations, multi-view images maintain high fidelity in appearance, as they capture real-world textures directly from images. However, they require sophisticated algorithms for accurate depth estimation and 3D reconstruction, and managing a large number of images can lead to high computational complexity.

\subsubsection{Explicit 3D Representations}

Explicit representations directly define the geometric shape of 3D models, providing clear and detailed structures. 
These representations are intuitive and easy to manipulate, but they often require significant storage space, especially for complex shapes.

\paragraph{Point Clouds} 

Point clouds \citep{scannet} represent 3D objects or scenes by sampling points in three-dimensional space. Each point in a point cloud is characterized by its spatial coordinates (x, y, z) and can optionally include additional attributes such as color, normal vectors, or intensity values. Point clouds are a direct output of many 3D scanning devices like LiDAR, structured light scanners, or stereo vision systems. 
Point clouds offer a sparse representation of a surface, capturing only the sampled points rather than continuous surfaces. While simple and efficient for storing raw geometric data, point clouds often lack the topological information needed to define object surfaces or relationships between points, making it harder to perform tasks like rendering or simulation without further processing. To enhance their usability, point clouds are often converted into more structured forms such as meshes or voxel grids.
Point cloud data is widely used in various applications, including autonomous driving (for environment perception), robotics (for navigation and mapping), and 3D reconstruction (for scanning objects or scenes).

\paragraph{Voxel Grids}

Voxel grids \citep{voxelnet} discretize 3D space into small, equally sized cubic units known as "voxels" (volume elements), which are analogous to pixels in 2D images. Each voxel represents a small portion of space and can hold properties like occupancy (whether the voxel contains part of an object), color, or density.
Voxel grids are particularly advantageous for applications requiring volumetric analysis or when it's necessary to explicitly model the interior and exterior of objects, such as medical imaging (e.g., CT or MRI scans), volumetric rendering, and certain physics simulations. One key advantage of voxel grids is their regular structure, which makes them amenable to direct use in 3D convolutional neural networks (3D CNNs) for tasks such as object classification, segmentation, and scene understanding.
However, voxel grids can be computationally expensive due to the cubic growth of data size with resolution, which can lead to high memory requirements for detailed objects or large scenes. To mitigate this, techniques like octrees, which hierarchically represent voxels at different resolutions, are often used to reduce the computational burden while preserving geometric details.

\paragraph{Meshes}

It \citep{shapenet,partnet} represent 3D surfaces using interconnected vertices, edges, and faces, typically forming polygons (most often triangles or quadrilaterals). Meshes are widely used in computer graphics, 3D modeling, and gaming due to their efficiency in representing complex surfaces with a relatively low number of data points. 
Each vertex in a mesh corresponds to a point in 3D space (x, y, z), and the edges define the connections between vertices, forming the structure of the object. The faces, typically composed of triangles or quads, form the surfaces that define the shape of the object. Meshes are especially efficient for rendering, as they allow for smooth and detailed surface representation while minimizing the amount of data needed compared to volumetric representations.
Meshes are not only used for visualization but also for simulations involving physical properties like deformations, collisions, and animations. Moreover, they are an integral part of 3D printing workflows, where the mesh defines the boundary of the object to be printed. Mesh simplification techniques can also be applied to reduce the complexity of a mesh while retaining important geometric features, making meshes versatile for both high-precision and real-time applications.

\subsubsection{Implicit 3D Representations}

Implicit representations define 3D structures through mathematical functions rather than direct geometry. 
These representations are compact and can efficiently capture complex shapes, but decoding them into explicit geometry can be more challenging.




\paragraph{Neural Implicit Representations}

Neural implicit representations \citep{nerf} are a powerful approach to encoding 3D geometry using neural networks. They represent shapes not through explicit vertices and faces, but as continuous functions that can be queried at any point in 3D space. This method allows for high levels of detail and smooth surface representations, making it particularly suitable for complex shapes. One of the key advantages of neural implicit representations is their ability to learn from various input data, such as 2D images, point clouds, or voxel grids, enabling them to capture intricate details and variations in geometry effectively.

\paragraph{Occupancy Fields}

An occupancy field \citep{occupancy_fields} is a specific type of neural implicit representation that determines whether a point in 3D space is occupied by a surface or not. This is typically modeled as a binary function where each point returns a value indicating occupancy (1) or non-occupancy (0). By training on 3D data, such as point clouds or volumetric data, occupancy fields can learn to approximate complex surfaces and provide a compact representation of the object. This representation is particularly useful in applications like scene understanding and object reconstruction, as it allows for efficient querying and rendering of surfaces.

\paragraph{Signed Distance Fields (SDFs)}

They \citep{signed_distance_fields} are a popular form of implicit representation that encode the geometry of 3D shapes based on the distance from a given point to the nearest surface. The SDF function returns negative values for points inside the object, zero at the surface, and positive values for points outside. This characteristic allows SDFs to provide smooth and continuous representations of surfaces, making them ideal for tasks like shape blending, deformation, and collision detection. SDFs are also beneficial in applications such as 3D shape generation and rendering, as they can be efficiently used in ray marching algorithms to visualize complex shapes.

\paragraph{3D Gaussian Splatting}

3D Gaussian splatting \citep{3dgs} is an emerging technique that utilizes Gaussian functions to represent 3D geometry in a probabilistic manner. In this approach, each point in space is associated with a Gaussian blob, which can be combined with other Gaussian representations to form a more complex shape. This method enables the modeling of intricate details and soft boundaries, providing a more organic representation of surfaces. 3D Gaussian splatting is particularly advantageous in scenarios where data is sparse or noisy, as it inherently incorporates uncertainty and smoothness. Applications of this technique include volumetric rendering, scene reconstruction, and enhancing neural rendering methods.

\subsection{Challenges in 3D Vision}

\paragraph{Occlusion} 
Occlusion is a major challenge in 3D vision as it arises when parts of an object are obscured by other objects in the scene. This issue is particularly problematic in densely cluttered environments, where multiple objects overlap or block each other. In such cases, 3D sensors like LiDAR or stereo cameras may not capture all surfaces of the occluded object, resulting in incomplete or distorted data. The missing information makes it difficult for downstream tasks such as object recognition, surface reconstruction, or scene understanding. Various techniques, such as multi-view aggregation, depth completion, and occlusion-aware models, have been developed to mitigate this issue, but occlusion remains a challenging problem, particularly in real-time applications like autonomous driving or robotic navigation, where dynamic objects can continuously block parts of the scene.

\paragraph{Varying Point Density}  
In 3D vision, the point clouds generated by scanning devices often suffer from varying point density. Some regions of a scanned object or scene may have dense sampling, while other regions may be sparsely populated with points. This variability is influenced by factors such as the distance from the sensor to the object, the angle of incidence of the scanning beam, and the reflective properties of surfaces. For example, objects close to the sensor or directly facing it may have dense point coverage, while those further away or at oblique angles may have fewer points. This non-uniformity can lead to challenges in 3D surface reconstruction, feature extraction, and object detection. Algorithms for point cloud up-sampling, surface interpolation, and non-uniform sampling compensation are used to address this issue, but the trade-off between computational complexity and real-time performance remains a concern, especially in large-scale environments.

\paragraph{Noise and Outliers}  
Noise and outliers are inherent challenges in 3D data acquisition, often caused by limitations in sensor precision or environmental factors. Noise refers to small deviations in the position of points due to inaccuracies in the scanning process, while outliers refer to points that are erroneously placed far from their true locations, possibly resulting from reflections, sensor calibration errors, or environmental interference. These imperfections can distort the shape and structure of 3D objects, leading to unreliable data for further processing. Techniques such as denoising filters, outlier removal algorithms, and robust statistical methods have been developed to handle these issues. However, finding the right balance between eliminating noise and preserving important details in the 3D data remains challenging, especially in applications like medical imaging or high-precision industrial inspections, where accuracy is critical.

%% file: Section/section4_tasks.tex
\section{3D Diffusion Generation}
\label{sec:tasks}

\input{Figures/fig_tree}

In this section, we provide an overview of various tasks involved in 3D diffusion generation, which take advantage of advanced machine learning techniques to create, manipulate, and analyze 3D data. These tasks are driven by the capability of diffusion models to learn complex data distributions, making them effective for generating and refining 3D content.

\input{Section/section4.1_uncon_gen}

\input{Section/section4.2_image_to_3d}

\input{Section/section4.3_text_to_3d}

\input{Section/section4.4_texture_gen}

\input{Section/section4.5_human_avatar}

\input{Section/section4.6_scene_gen}

\input{Section/section4.7_3d_editing}

\input{Section/section4.8_novel_view}

\input{Section/section4.9_depth}



%% file: Figures/fig_tree.tex
\tikzstyle{my-box}=[
    rectangle,
    draw=hidden-draw,
    rounded corners,
    text opacity=1,
    minimum height=1.5em,
    minimum width=5em,
    inner sep=2pt,
    align=center,
    fill opacity=.5,
    line width=0.8pt,
]
\tikzstyle{leaf}=[my-box, minimum height=1.5em,
    fill=hidden-pink!80, text=black, align=left,font=\tiny,
    inner xsep=2pt,
    inner ysep=4pt,
    line width=0.8pt,
]
\begin{figure*}
    \centering
    \resizebox{0.9\textwidth}{!}{
        \begin{forest}
            forked edges,
            for tree={
                grow=east,
                reversed=true,
                anchor=base west,
                parent anchor=east,
                child anchor=west,
                base=left,
                font=\footnotesize,
                anchor=center,
                align=center, 
                text centered,
                rectangle,
                draw=hidden-draw,
                rounded corners,
                align=left,
                minimum width=2em,
                edge+={darkgray, line width=1pt},
                s sep=3pt,
                inner xsep=2pt,
                inner ysep=3pt,
                line width=0.8pt,
                ver/.style={rotate=90, child anchor=north, parent anchor=south, anchor=center, font=\normalsize\bfseries},
            },
            where level=1{text width=3.5em,font=\footnotesize,}{},
            where level=2{text width=4.5em,font=\footnotesize,}{},
            where level=3{text width=3.0em,font=\footnotesize,}{},
            [
                3D Diffusion task, ver
                [
                    Unconditional \\ Generation, text width= 6em, anchor=center, align=center, text centered, font=\small
                    [
                        Explicit, text width= 3.3em, anchor=center, align=center, font=\small
                        [
                            PDR \citep{pdr}{,} RenderDiffusion \citep{renderdiffusion}{,}
                            Point-Voxel Diffusion \citep{point_diffusion}{,}  \\
                            TIGER \citep{tiger}{,}
                            PointDif \citep{point_pretrain}{,}
                            DiT-3D \citep{dit3d}{,}  LION \citep{lion}{,} \\
                            SDS-Complete \citep{point_comp_diffu}{,} 
                            Nunes \etal \citep{scaling_diffusion}{,} 
                            3DShape2VecSet \citep{3dshape2vecset}
                            , leaf, text width=46em, anchor=center, align=center, text centered, font = \scriptsize
                        ]
                    ]
                    [
                        Implicit, text width= 3.3em, anchor=center, align=center, font=\small
                        [    
                            SDF-Diffusion \citep{diffusion_sdf_3d}{,}
                            Diffusion-SDF \citep{diffu_sdf}{,}
                            DiffRF \citep{diffrf}{,} \\
                            Triplane Diffusion \citep{3dnerf_diff}{,}
                            3D-LDM \citep{3dldm}
                            , leaf, text width=46em, anchor=center, align=center, text centered, font = \scriptsize
                        ]
                    ]
                ]
                [
                    Image to 3D, text width= 6em, anchor=center, align=center, font=\small
                    [
                        EfficientDreamer \citep{efficientdreamer}{,}
                        Wonder3d \citep{wonder3d}{,}
                        InstantMesh \citep{instantmesh}{,} Lgm \citep{lgm}{,} \\ SJC \citep{score_jacobian_chaining}{,} GRM \citep{grm} Magic123 \citep{magic123}{,} One-2-3-45 \citep{one2345}{,} \\
                        Neurallift-360 \citep{neurallift}{,}
                        Viewset Diffusion \citep{viewset_diffusion}{,} 
                        MeshDiffusion \citep{meshdiffusion}{,}
                        \\ RealFusion \citep{realfusion}{,}
                        Make-It-3D \citep{make_it_3d}{,} 
                        LAS-Diffusion \citep{locally_sdf}{,} \\
                        Consistent123 \citep{consistent123}
                        , leaf, text width=51em, anchor=center, align=center, text centered, font = \scriptsize
                    ]
                ]
                [
                    Text to 3D, text width= 6em, anchor=center, align=center, font=\small
                    [
                        Point-e \citep{pointe}{,} GSGEN \citep{text_3dgs}{,}
                        Latent-NeRF \citep{latent_nerf}{,} 
                        3DFuse \citep{seolet}{,} \\
                        Fantasia3D \citep{fantasia3d}{,}
                        STPD \citep{sketch_text}{,} 
                        GSGEN \citep{text_3dgs}{,} 
                        AYG \citep{align_3dgs}{,}  \\
                        Direct2.5 \citep{direct25}{,} DreamFusion \citep{dreamfusion}{,} 
                        4d-fy \citep{4dfy}{,} 
                        Dreamtime \citep{dreamtime}{,} \\ GaussianDreamer \citep{gaussiandreamer}{,}  Diffusion-SDF \citep{diffu_sdf_text}{,} ProlificDreamer \citep{prolificdreamer}
                        , leaf, text width=51em, anchor=center, align=center, text centered, font = \scriptsize
                    ]
                ]
                [
                    Texture \\ Generation, text width= 5.6em, anchor=center, align=center, font=\small
                    [
                        TexFusion \citep{texfusion}{,}
                        Point-UV Diffusion \citep{point_uv_diffu}{,}
                        3DGen \citep{3dgen}{,} 
                        TexOct \citep{uvidm}{,} \\
                        Paint3D \citep{paint3d}{,}
                        CRM \citep{crm}{,} 
                        TexPainter \citep{texpainter}{,} SceneTex \citep{scenetex}{,} \\
                        TEXTure \citep{texture_3d_shape}{,}
                        DreamMat \citep{dreammat}{,} 
                        Paint3D \citep{paint3d}{,} \\
                        Text2Tex \citep{text2tex}{,}  One-2-3-45++ \citep{one2345++}{,} 
                         TextureDreamer \citep{texturedreamer} 
                        , leaf, text width=51em, anchor=center, align=center, text centered, font = \scriptsize
                    ]
                ]
                [
                    Human Avatar \\ Generation, text width= 6.4em, anchor=center, align=center ,font=\small
                    [
                        Avatar, text width=3em, , anchor=center, align=center, font=\small
                        [
                            DiffusionGAN3D \citep{diffusiongan3d}{,} RODIN \citep{rodin}{,}
                            AvatarCraft \citep{avatarcraft}{,} \\
                            Chen \etal \citep{morphable}{,} 
                            UltrAvatar \citep{ultravatar}{,} 
                            UV-IDM \citep{uvidm}
                            , leaf, text width=45.8em, anchor=center, align=center, text centered, font = \scriptsize
                        ]
                    ]
                    [
                        Pose, text width= 3em, anchor=center, align=center, font=\small
                        [
                            PrimDiffusion \citep{primdiffusion}{,}
                            Chupa \citep{chupa}{,}
                            DINAR \citep{dinar}{,} 
                            SiTH \citep{sith}{,} \\
                            ScoreHMR \citep{score_human_rec}{,}
                            HumanNorm \citep{humannorm}{,} 
                            DreamHuman \citep{dreamhuman}
                            , leaf, text width=45.8em, anchor=center, align=center, text centered, font = \scriptsize
                        ]
                    ]
                    [
                        Action, text width= 3em, anchor=center, align=center, font=\small
                        [
                            AnimateMe \citep{animateme}{,}
                            InterDiff \citep{interdiff}{,}
                            PhysDiff \citep{physdiff}
                            , leaf, text width=45.8em, anchor=center, align=center, text centered, font = \scriptsize
                        ]
                    ]
                ]
                [
                    Scene \\ Generation, text width= 6em, anchor=center, align=center, font=\small
                    [
                        Image-guided, text width= 5.7em, anchor=center, align=center, font=\small
                        [
                            Kim \etal \citep{neuralfield}{,}
                            Wynn \etal \citep{diffusionerf}{,} LiDMs \citep{scene_lidar_diffu} {,} \\
                            LucidDreamer \citep{luciddreamer}{,}
                            BlockFusion \citep{blockfusion}{,}
                            , leaf, text width=43.5em, anchor=center, align=center, text centered, font = \scriptsize
                        ]
                    ]
                    [
                        Text-guided, text width= 5.6em, anchor=center, align=center, font=\small
                        [
                            SceneScape \citep{scenescape}{,}
                            Po \etal \citep{compositional_3d}{,}
                            Text2NeRF \citep{text2nerf}
                            , leaf, text width=43.5em, anchor=center, align=center, text centered, font = \scriptsize
                        ]
                    ]
                    [
                        Scene graph \\ guided, text width= 5.7em , anchor=center, align=center, font=\small
                        [
                            GraphDreamer \citep{graphdreamer}{,}
                            DiffuScene \citep{diffuscene}{,}
                            EchoScene \citep{echoscene}
                            , leaf, text width=43.5em, anchor=center, align=center, text centered, font = \scriptsize
                        ]
                    ]
                ]
                [
                    3D Editing \\ \& \\ Manipulation, text width= 6em, anchor=center, align=center, font=\small
                    [
                        DATID-3D \citep{datid3d}{,}
                        SKED \citep{sked}{,}
                        Vox-E \citep{voxe}{,} 
                        HeadSculpt \citep{headsculpt}{,} \\
                        DreamEditor \citep{dreameditor}{,}
                        SketchDream \citep{sketchdream}{,} 
                        GaussianEditor \citep{gaussianeditor}{,} \\
                        3D Paintbrush \citep{3d_paintbrush}{,} 
                        Pandey \etal \citep{diffusion_handles}{,} 
                        APAP \citep{plausible}
                        , leaf, text width=50.5em, anchor=center, align=center, text centered, font = \scriptsize
                    ]
                ]
                [
                    Novel View \\ Synthesis, text width= 6em, anchor=center, align=center, font=\small
                    [
                        NeRDi \citep{nerdi}{,} 
                        HOLODIFFUSION \citep{holodiffusion}{,}
                        Tseng \etal \citep{consistent_view}{,} \\
                        SparseFusion \citep{sparsefusion}{,}
                        Zero-1-to-3 \citep{zero1to3}{,}
                        Chan \etal \citep{generative_novel_view}{,} \\
                        Zero123++ \citep{zero123++}{,} 
                        DiM \citep{nvs_diffusion}{,} 
                        NerfDiff \citep{nerfdiff}{,} 
                        ViewDiff \citep{viewdiff}{,}  \\
                        Consistent-1-to-3 \citep{consistent1to3}{,} 
                        SyncDreamer \citep{syncdreamer}
                        , leaf, text width=50.5em, anchor=center, align=center, text centered, font = \scriptsize
                    ]
                ]
                [
                    Depth \\Estimation, text width= 6em, anchor=center, align=center, font=\small
                    [
                        DDP \citep{ddp_dense}{,} 
                        DADP \citep{diff_aug_depth}{,} 
                        DiffusionDepth \citep{diffusiondepth}{,}
                        Atlantis \citep{atlantis}{,} \\
                        Tosi \etal \citep{diffusion_challenge}{,}
                        SEDiff \citep{sediff}{,}
                        ECoDepth \citep{ecodepth}{,} 
                        DDVM \citep{surprising}{,} \\
                        Marigold \etal \citep{repurposing}
                        , leaf, text width=50.5em, anchor=center, align=center, text centered, font = \scriptsize
                    ]
                ]
            ]
        \end{forest}
    }
    \caption{Taxonomy of 3D Diffusion Tasks. . }
    \label{taxo_of_videogeneration}
\end{figure*}

%% file: Section/section4.1_uncon_gen.tex
\subsection{Unconditional Generation} 
\label{sec:uncon_gen}

\begin{figure}[H]
  \centering
  \includegraphics[width=\linewidth]{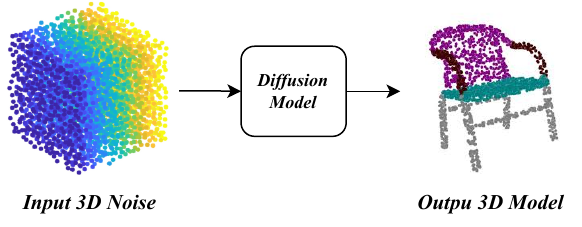}
  \caption{Example of unconditional 3D generation task.}
  \label{fig:uncon_gen}
\end{figure}

Unconditional 3D generation focuses on autonomously synthesizing three-dimensional structures without relying on explicit inputs such as class labels, images, or text prompts, as shown in Figure~\ref{fig:uncon_gen}. 
By learning patterns from training data, models generate diverse 3D outputs, such as shapes, scenes, or point clouds, directly from random noise, aiming to capture the underlying data distribution for realistic and varied results. 
Early breakthroughs include \citet{renderdiffusion}, who achieved 3D generation using 2D diffusion models by rendering intermediate 3D representations at each denoising step, enforcing 3D consistency through inductive biases despite relying on 2D supervision. 
Building on this, \citet{point_diffusion} proposed Point-Voxel Diffusion, a probabilistic framework that reverses the diffusion process to transform noisy point clouds into structured 3D shapes using point-voxel hybrid representations. 
Subsequent work by \citet{tiger} revealed the dynamic roles of attention and convolution in diffusion: global attention mechanisms dominate early stages to define overall shapes, while local convolutions refine surface details in later stages. 
Their time-varying denoising model adaptively fuses these features via optimizable masks. 
Pre-training strategies such as PointDif \citep{point_pretrain} further enhance robustness by defining point cloud recovery as a conditional generation task with recurrent uniform sampling. Hybrid approaches such as \citet{dit3d} bridge the 2D and 3D domains by voxelizing point clouds and fine-tuning pre-trained 2D models for high-quality 3D synthesis.  

Implicit representations have broadened 3D generation capabilities. Methods like \citet{diffusion_sdf_3d} leverage Signed Distance Functions (SDFs), first generating low-resolution SDFs and then applying diffusion-based super-resolution to refine details, enabling direct mesh extraction via Marching Cubes. Similarly, \citet{diffu_sdf} modulate the SDFs into diffusion-trained latent vectors, reconstructing shapes from denoised vectors. Radiance fields also benefit from diffusion: \citet{diffrf} generate explicit voxel-grid RFs using 3D DDPMs, while \citet{3dnerf_diff} project 3D scenes into 2D triplanes, leveraging 2D diffusion models to synthesize neural RFs efficiently.  

Latent space compression strategies improve both quality and efficiency. For instance, \citet{3dldm} train auto-decoders on SDFs to construct a compact latent space, where diffusion models generate continuous 3D surfaces. \citet{lion} propose hierarchical VAEs with dual diffusion models, where one for global shape vectors and another for point cloud structures, enabling fine-grained control. \citet{3dshape2vecset} introduce transformer-friendly latent representations using Radial Basis Functions and attention mechanisms, supporting diffusion-based generation of neural fields.  

Beyond pure generation, diffusion models excel in 3D completion. \citet{pdr} refine incomplete point clouds through a two-stage coarse-to-fine diffusion process, first generating a rough shape and then improving the details. \citet{point_comp_diffu} integrate semantic guidance from text-to-image diffusion models to complete surfaces using textual cues from partial input. For LiDAR data, \citet{scaling_diffusion} redefine the noise application by perturbing individual points locally rather than globally, preserving scene structure while learning intricate details. These advancements underscore the adaptability of DDPMs in both the creation and refining of 3D data in applications such as virtual environments, gaming, and autonomous systems.

%% file: Section/section4.2_image_to_3d.tex
\subsection{Image-to-3D Generation} 
\label{sec:img_to_3d}

\begin{figure}[h]
  \centering
  \includegraphics[width=\linewidth]{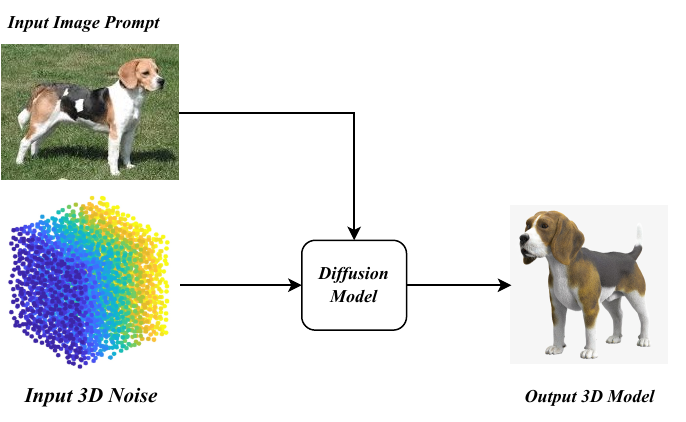}
  \caption{Example of image to 3D generation task.}
  \label{fig:image_to_3d}
\end{figure}

Image-to-3D generation, as shown in Figure~\ref{fig:image_to_3d}, involves converting 2D images into 3D representations by leveraging diffusion models to infer depth, structure, and texture from limited views. This task is critical for applications such as 3D reconstruction and enhancing single-view predictions via iterative refinement of spatial understanding. Diffusion models act as powerful generative priors to resolve the ambiguities inherent in reconstructing 3D objects from sparse 2D supervision. Key innovations include integrating differentiable rendering into denoising processes and synthesizing novel views using pre-trained 2D diffusion models.

Recent advances address the challenge of single-image 3D reconstruction by combining diffusion priors with geometric constraints. For example, SJC~\citep{score_jacobian_chaining} chains gradients from a diffusion model with a differentiable Jacobian renderer matrix to translate 2D image gradients into 3D asset updates, allowing 3D generation without 3D training data. 
And NeuralLift \citep{neurallift} introduces CLIP-guided sampling to merge diffusion priors with reference images, achieving realistic 3D reconstructions without precise depth estimation. Their scale-invariant depth supervision further reduces reliance on multi-view depth consistency. 
Furthermore, Viewset Diffusion \citep{viewset_diffusion} trains a diffusion model on multi-view 2D data and embeds a bijective mapping between viewsets and 3D models within the denoising network, ensuring geometric alignment. 
And MeshDiffusion \citep{meshdiffusion} employs deformable tetrahedral grids to represent 3D meshes, bypassing topology irregularities, and trains a score-based diffusion model directly on parametric mesh representations for effective generation.

When reconstructing 3D models from single-view images, missing backside information is mitigated by generating plausible multi-view hypotheses. 
For example, RealFusion \citep{realfusion} optimizes text prompts to guide diffusion models in synthesizing novel views, enabling unsupervised 3D reconstruction of arbitrary objects. 
And One-2-3-45 \citep{one2345} generates multi-view images via 2D diffusion and applies cost-volume-based neural surface reconstruction, while EfficientDreamer \citep{efficientdreamer} produces four orthogonal views from text prompts to impose geometric priors. 
Furthermore, Wonder3D \citep{wonder3d} enhances multi-view consistency through cross-domain attention and geometry-aware normal fusion. 
And InstantMesh \citep{instantmesh} accelerates high-quality mesh generation by combining multi-view diffusion with a sparse view reconstruction framework based on the Large Reconstruction Model, directly predicting meshes from consistent multi-view inputs.

Two-stage approaches, though less efficient than end-to-end methods, remain prevalent for their accuracy. Make-It-3D \citep{make_it_3d} first optimizes NeRF using diffusion priors and reference image constraints, then refines the model into a textured point cloud with high-fidelity textures. 
LAS-Diffusion \citep{locally_sdf} employs a coarse-to-fine strategy: an occupancy-diffusion stage generates low-resolution shape shells, followed by SDF-diffusion to refine geometric details. It introduces view-aware local attention to align 2D image patches with 3D voxel features. 
On the other hand, Consistent123 \citep{consistent123} combines 3D structural priors with CLIP-guided adaptive detection in its first stage, then prioritizes 2D texture priors to enhance details. 
And Magic123 \citep{magic123} transitions from NeRF-based coarse geometry to memory-efficient mesh optimization, balancing exploration (via 2D diffusion) and exploitation (via 3D priors) during generation.

Recent trends explore 3D Gaussian splatting \citep{3dgs} as an alternative to NeRF for efficient rendering. For example, LGM \citep{lgm} introduces multi-view Gaussian features for differentiable rendering and integrates them with an asymmetric U-Net to generate 3D models rapidly from single images. 
And GRM \citep{grm} reconstructs 3D assets from sparse views using a transformer-based architecture that maps pixels to 3D Gaussian parameters, which are re-projected to form dense scene representations. These methods highlight the growing synergy between diffusion models and modern 3D representations to advance reconstruction speed, quality, and scalability.

%% file: Section/section4.3_text_to_3d.tex
\subsection{Text-to-3D Generation} 
\label{sec:text_3d}

\begin{figure}[h]
  \centering
  \includegraphics[width=\linewidth]{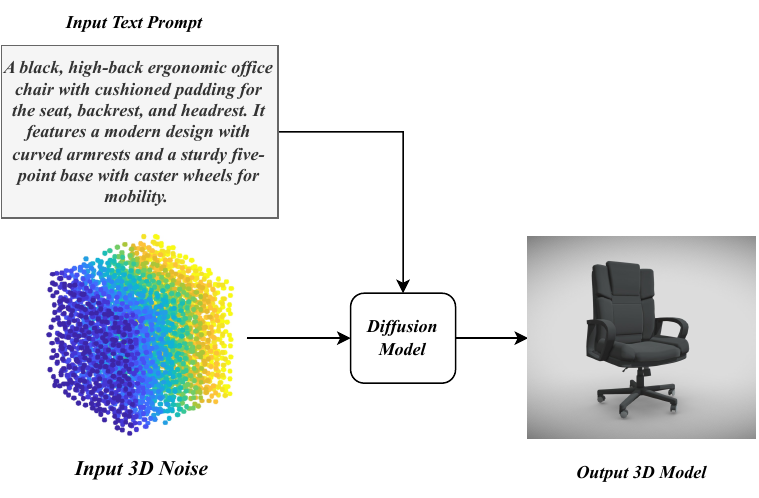}
  \caption{Example of text to 3d generation task.}
  \label{fig:text_to_3d}
\end{figure}

Text-to-3D generation focuses on converting textual descriptions into 3D models or scenes, as shown in Figure~\ref{fig:text_to_3d}. 
Diffusion models have become central to this task due to their ability to connect language semantics with 3D structural representations, enabling the creation of meaningful assets for applications like gaming, virtual environments, and industrial design. Early advances include the work of \citet{pointe}, who pioneered a dual-stage process: a text-to-image diffusion model first synthesizes a single view 2D representation, which then guides a secondary conditional diffusion model to generate a 3D point cloud. Building on this, \citet{diffu_sdf_text} developed a patch-based autoencoder to encode truncated signed distance fields of 3D shapes into localized Gaussian latent spaces. Their approach integrates a voxelized diffusion model to model both inter-patch relationships and global structural coherence. 
To streamline computational demands, \citet{latent_nerf} introduced Latent-NeRF, which bypasses the RGB space by operating entirely in a compressed latent space, reducing the need for repeated image encoding during guidance steps. Addressing the inherent 3D inconsistencies of 2D diffusion models, \citet{seolet} proposed generating preliminary 3D point clouds from text, projecting them into multi-view depth maps to enforce explicit 3D constraints. 
These depth maps are then fed into the diffusion process, enabling viewpoint-aware optimization for coherent scene generation.

For higher precision, many two-stage frameworks have been proposed. For example, \citet{fantasia3d} separates geometry and appearance modeling, using text features to independently guide shape formation and surface properties. Similarly, \citet{sketch_text} employs a sequential pipeline: a diffusion model first generates a geometric sketch, followed by colorization based on the established shape, ensuring alignment between structure and texture.

Recent work utilizing 3D Gaussian splatting demonstrates further innovation. \citet{text_3dgs} introduces GSGEN, which combines 2D and 3D diffusion priors in a two-phase optimization: first, constructing a coarse 3D structure and then refining details via a compactness-driven densification process. For dynamic scenes, \citet{align_3dgs} combines dynamic 3D Gaussian mixtures with deformation fields to represent 4D motion, integrating text-to-image, text-to-video, and multi-view diffusion models within a score distillation framework. Meanwhile, \citet{gaussiandreamer} bridges 3D initialization and 2D refinement through Gaussian splatting in their GaussianDreamer framework, where a 3D diffusion prior seeds the geometry, and the 2D diffusion iteratively improves both shape and appearance.

Enhancements in score distillation sampling (SDS) address optimization challenges. For example, \citet{prolificdreamer} reimagines SDS as a variational process, treating 3D parameters as probabilistic distributions to align multi-view renderings with 2D diffusion priors. And \citet{4dfy} adopts a hybrid strategy, alternating between static scene initialization and variational SDS-driven appearance refinement to mitigate artifacts, such as the ``Janus problem''. Recognizing inefficiencies in uniform timestep sampling, \citet{dreamtime} proposes Time-Prioritized SDS, using a decay function to prioritize critical denoising steps during 3D optimization, accelerating convergence, and improving output quality.

To avoid the scarcity of 3D training data, methods such as \citet{dreamfusion} bypass 3D supervision entirely by optimizing NeRF representations through 2D diffusion-guided loss functions, distilling 3D consistency from multi-view image projections. 
Expanding on this, \citet{direct25} fine-tunes pre-trained 2D diffusion models into a multi-view 2.5D framework, directly capturing 3D structural distributions while preserving the diversity and generalization of 2D models. 
In summary, these approaches demonstrate that high-fidelity 3D generation can thrive even without direct 3D data, opening new avenues for scalable content creation.

%% file: Section/section4.4_texture_gen.tex
\subsection{Texture Generation} 
\label{sec:texture_gen}

\begin{figure}[h]
  \centering
  \includegraphics[width=\linewidth]{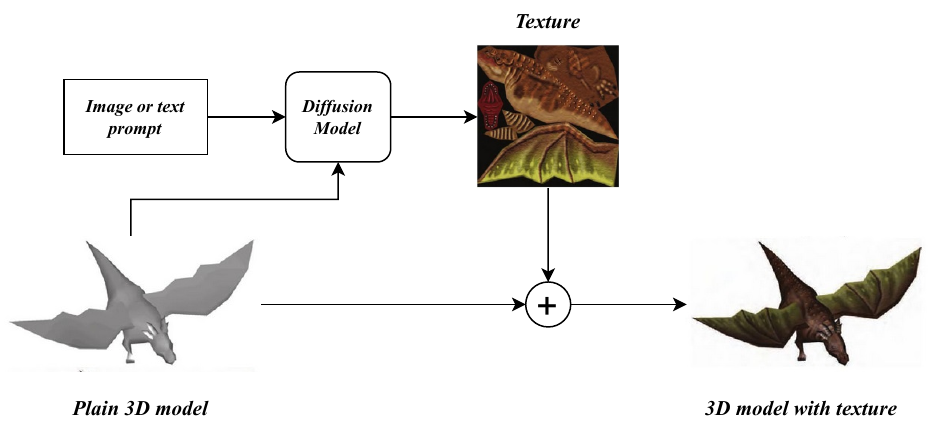}
  \caption{Example of texture generation task.}
  \label{fig:texture_gen}
\end{figure}

Texture generation focuses on mapping realistic surface details onto 3D meshes, with diffusion models playing a key role in synthesizing high-fidelity textures that preserve geometric accuracy and surface realism, as shown in Figure~\ref{fig:texture_gen}. These methods enhance applications in graphics, simulations, and AR and VR by aligning generated textures with real-world appearances.

Specifically, \citet{texfusion} propose TexFusion, a text-guided 3D texture synthesis framework that employs a latent diffusion model and a 3D-consistent sampling strategy. By iteratively denoising 2D rendered views and aggregating predictions into a unified latent texture map, the method produces detailed, text-aligned textures while maintaining 3D coherence. 
And \citet{point_uv_diffu} address the geometry compatibility through Point-UV diffusion, a two-stage coarse-to-fine framework that combines UV mapping with diffusion models for seamless texture generation. 
Then, \citet{3dgen} leverage a hybrid approach, integrating a Triplane VAE with a conditional diffusion model to reconstruct textured meshes from encoded point clouds in a triplane Gaussian latent space. 
For direct 3D optimization, \citet{uvidm} designed TexOct, an octree-based diffusion model that efficiently represents surface-sampled point clouds, resolving occlusion and sparse sampling artifacts. 
Meanwhile, \citet{texturedreamer} extract textures from limited images using personalized modeling and geometry-aware score distillation, \citet{crm} accelerate single-image mesh generation via a convolution and DDPM-based reconstruction model with geometric priors.

Multi-view consistency remains a critical challenge. 
For example, \citet{texpainter} ensure coherence by decoding noise-free latent states from a pre-trained diffusion model (e.g., DDIM) into color space at each denoising step, followed by latent code optimization to align rendered views. 
And \citet{one2345++} adopt a two-phase method: they first fine-tune a 2D diffusion model to generate six consistent viewpoints, then apply a coarse-to-fine 3D diffusion model to predict textured meshes from these generated views. Lightweight optimization further refines the texture quality. 

Lighting integration also influences texture realism. \citet{dreammat} train a diffusion model under specific lighting conditions to produce shadow-free PBR materials, ensuring consistency with geometry. In contrast, \citet{paint3d} prioritize lighting-agnostic textures using a pre-trained 2D diffusion model and a specialized UV completion-upscaling pipeline for text-guided and image-guided generation.

Depth-aware methods improve visual consistency. 
For example, \citet{text2tex} utilize a depth-aware diffusion model to synthesize high-resolution partial textures from text prompts, progressively refining local textures via viewpoints to avoid artifacts. And \citet{texture_3d_shape} employ depth-to-image diffusion with dynamic trimap segmentation for seamless multi-view textures. 
Then, \citet{scenetex} re-imagine texture synthesis as an RGB-space optimization task, combining multi-resolution texture fields and cross-attention decoders to leverage depth-to-image priors for stylistically coherent indoor scenes.

%% file: Section/section4.5_human_avatar.tex
\subsection{Human Avatar Generation} 
\label{sec:avatar}

\begin{figure}[h]
  \centering
  \includegraphics[width=\linewidth]{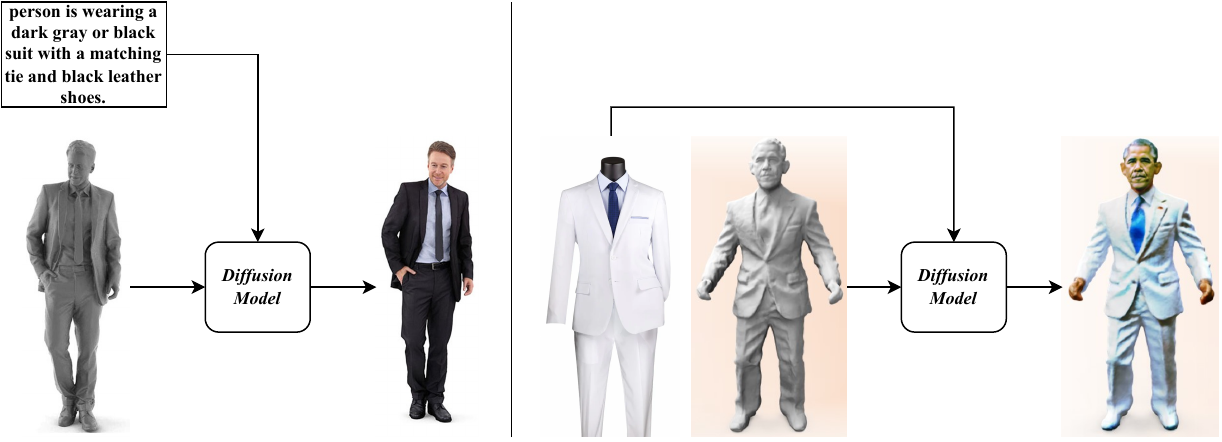}
  \caption{Example of 3D human avatar generation task.}
  \label{fig:avatar}
\end{figure}

\begin{figure*}[t]
  \centering
  \includegraphics[width=\linewidth]{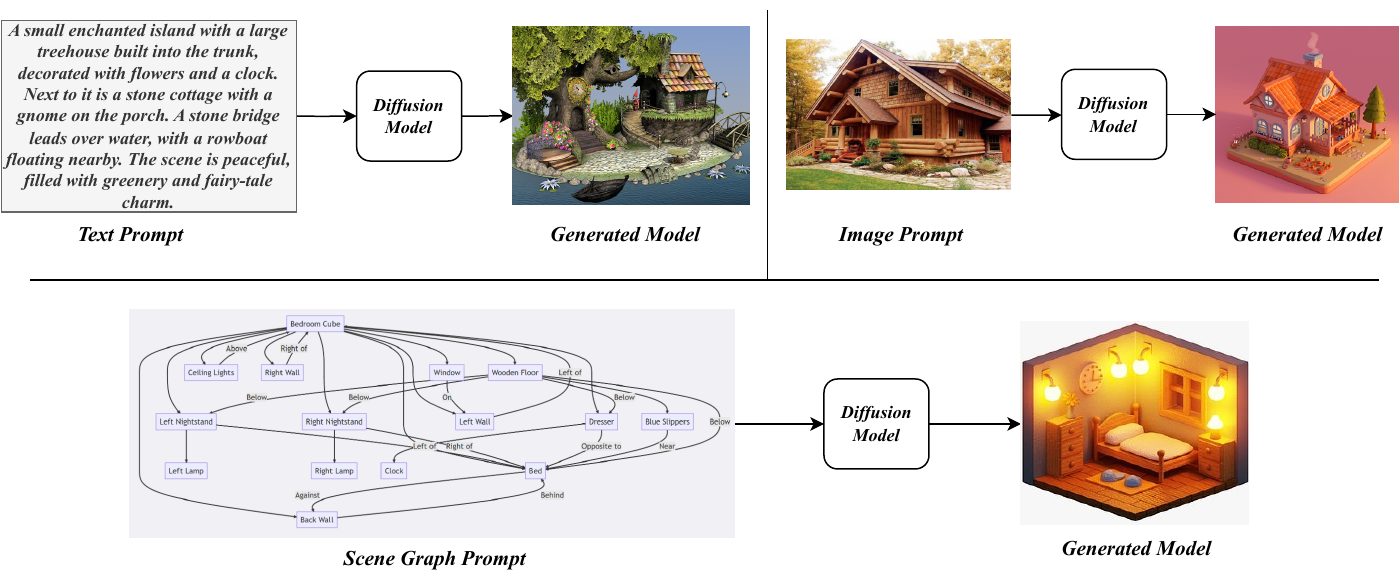}
  \caption{Example of 3D scene generation task.}
  \label{fig:scene_gen}
\end{figure*}

The generation of human avatars focuses on creating realistic or stylized 3D models of humans, as shown in Figure~\ref{fig:avatar}. Diffusion models improve anatomical accuracy, dynamic poses, and textures, enabling fine-tuning of facial expressions and body movements for applications in gaming, virtual reality, and digital fashion.

Current research prioritizes human head generation due to its technical focus and practical demand. For example, \citet{rodin} generates detailed 3D avatars using Neural Radiance Fields by projecting 3D feature maps onto 2D planes with 3D-aware convolutions, preserving spatial relationships during rendering. And \citet{avatarcraft} leverages text-guided diffusion models to produce avatars with parameterized control over body shape and pose. 
Meanwhile, \citet{morphable} introduces a deformation diffusion model that synthesizes 3D meshes conditioned on identity and expression parameters, further improved by multi-view consistency constraints for realistic outputs. The DiffusionGAN3D framework by \citet{diffusiongan3d} integrates pre-trained 3D generators with text-to-image diffusion models, enhancing diversity through Relative Distance Loss and adaptive triplanes. 
For animatable avatars, \citet{ultravatar} proposes UltrAvatar, combining texture diffusion and authenticity guidance to generate models from text or single images. To address the limitations of texture aspect, \citet{uvidm} employs latent diffusion models with the Basel Face Model to preserve identity and resolve occlusion issues in facial texture synthesis.

Generating full-body 3D models with specific poses introduces greater complexity. For example, \citet{primdiffusion} pioneered a diffusion-based framework using volumetric primitives, small 3D units encoding radiance and motion, to streamline computation and handle intricate body topology. 
And \citet{chupa} simplifies the process by decomposing it into 2D normal map generation and 3D reconstruction, ensuring multi-view consistency while reducing computational overhead. For stylized characters, \citet{dinar} combines neural textures with the parametric model SMPL-X, enabling cartoon-style avatars with efficient texture recovery through latent diffusion. 
The single-view reconstruction challenges are tackled by \citet{sith}, which predicts occluded back views using diffusion models before reconstructing complete meshes. 
On the other hand, \citet{score_human_rec} adopts a diffusion-based approach to mimic the fitting of the model, inferring human parameters from images through task-guided denoising. And \citet{humannorm} improves geometric accuracy through normal-aligned diffusion and progressive generation strategies. 
Unlike image-based methods, \citet{dreamhuman} generates animatable avatars directly from text by optimizing NeRF representations with instance-specific deformations.

Beyond static models, recent work explores dynamic motion synthesis. \citet{animateme} ensures temporal coherence in facial animations by applying graph neural networks as denoising diffusion models in mesh space. For human-object interactions, \citet{interdiff} combines interaction diffusion with physics-based corrections to generate long-term 3D sequences. Recognizing the limitations of traditional motion models, \citet{physdiff} integrates physics simulators into the diffusion pipeline, projecting denoised motions onto physically plausible trajectories through motion imitation. These advances highlight the expanding role of diffusion models in bridging realism, controllability, and computational efficiency across avatar generation and animation pipelines.

%% file: Section/section4.6_scene_gen.tex
\subsection{Scene Generation} 
\label{sec:scene_gen}

Scene generation involves building comprehensive 3D environments with objects, textures, and spatial arrangements. Diffusion models contribute significantly by synthesizing scenes with realistic object relationships, lighting, and textures, supporting applications like virtual worlds, architecture, and gaming. Common approaches include image-guided, text-guided, and scene graph-guided methods, as shown in Figure \ref{fig:scene_gen}.

Image-guided methods provide straightforward scene synthesis. 
For example, \citet{neuralfield} employ a scene auto-encoder to encode image pose pairs into neural fields, followed by latent compression and hierarchical diffusion modeling. 
To address geometry limitations in sparse-view NeRF scenarios, \citet{diffusionerf} integrate a DDPM trained on synthetic RGBD patches to regularize scene priors. And \citet{luciddreamer} develop LucidDreamer, iteratively refining 3D scenes through ``Dreaming'' (geometric image generation and 3D projection) and ``Alignment'' (point cloud integration). 
Then, \citet{blockfusion} propose BlockFusion, compressing 3D planes into latent space via VAE for scalable diffusion-based scene synthesis. 
On the other hand, for sensor data realism, \citet{scene_lidar_diffu} introduce LiDAR Diffusion Models that enhance geometric authenticity using prior knowledge.

Text-guided approaches enable for more flexible creative control. For example, \citet{scenescape} combine text-to-image models with depth prediction for 3D-consistent video generation. And \citet{compositional_3d} achieve localized scene control through bounding boxes and text prompts using Locally Conditioned Diffusion. Furthermore, \citet{text2nerf} integrate text-to-image diffusion with NeRF, employing progressive refinement and depth alignment for photorealistic multi-view scenes.

Scene graph-based methods leverage structured semantic relationships for logical scene hierarchies. For example, \citet{graphdreamer} pioneer GraphDreamer to resolve attribute conflicts in text-to-3D conversion using graph information. And \citet{diffuscene} design DiffuScene, a diffusion model that synthesizes unordered object collections with denoised attributes. Then, \citet{echoscene} develop EchoScene’s dual-branch architecture, where graph-convolutional information exchange during denoising ensures global consistency between the graph nodes of the scene. These approaches demonstrate how structured guidance improves spatial reasoning compared to unstructured image or text input.

%% file: Section/section4.7_3d_editing.tex
\subsection{3D Editing and Manipulation} 
\label{sec:3d_editing}

\begin{figure}[h]
  \centering
  \includegraphics[width=\linewidth]{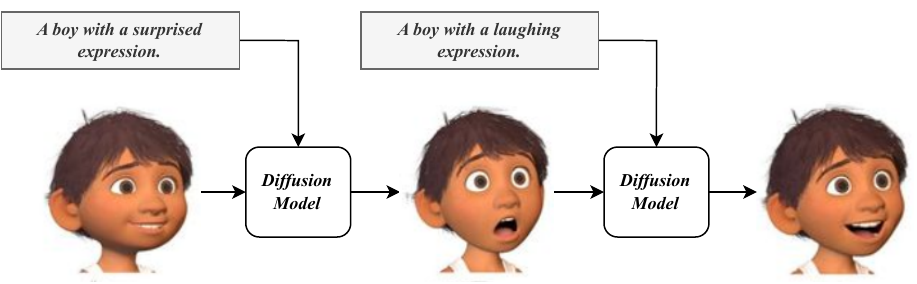}
  \caption{Example of 3D editing task.}
  \label{fig:3d_editing}
\end{figure}

Recent advances in 3D editing and manipulation leverage diffusion models and neural representations to achieve precise modifications while preserving structural integrity, as shown in Figure~\ref{fig:3d_editing}. 
These techniques primarily operate through direct manipulation of 3D data or latent representations, allowing efficient adjustments to shape, texture, and pose with minimal distortion. Several innovative approaches show this paradigm via distinct methodologies.

Diffusion-based domain adaptation methods~\citep{datid3d} utilize text-to-image diffusion models combined with adversarial training to transform samples from pre-trained 3D generators into diversified target images while maintaining text prompt diversity. 
Furthermore, multi-modal interaction techniques, such as SKED \citep{sked} and SketchDream \citep{sketchdream} integrate sketch guidance with text prompts, employing geometric reasoning for coarse positioning and generative models for detail refinement. These methods utilize multi-view sketches and depth-aware diffusion models to establish spatial correspondences and ensure 3D consistency via specialized loss functions.

Volumetric editing methods demonstrate progress in structural preservation. Vox-E \citep{voxe} introduces volumetric regularization losses that operate directly in a 3D space to maintain global structural correlations during voxel-based editing. For neural field representations, DreamEditor \citep{dreameditor} achieves localized editing through grid-based neural fields and semantic-aware diffusion models, while GaussianEditor \citep{gaussianeditor} enhances control via Gaussian semantic tracing and hierarchical splatting constraints.

Identity-preserving editing receives particular attention in avatar manipulation. HeadSculpt For example, \citep{headsculpt} establishes a coarse-to-fine workflow combining landmark-based ControlNet and text inversion for 3D-aware generation, followed by score-blending techniques to maintain facial identity during texture optimization. Similarly with previous methods, 3D Paintbrush \citep{3d_paintbrush} enables localized stylization via neural texture maps optimized through cascaded score distillation, demonstrating precise adherence to surface geometry.

Practical editing frameworks address real-world applicability constraints. For example, \citet{diffusion_handles} achieves training-free 3D manipulation through proxy depth-based activation lifting and 3D transformation projection. 
And APAP \citep{plausible} improves the realism of mesh deformation through differentiable Poisson solvers and SDS-optimized priors, employing LoRA-adapted diffusion models to preserve object identity during interactive editing.

These developments collectively advance 3D editing capabilities through improved geometric awareness, semantic control, and structural preservation mechanisms, significantly expanding practical applications in content creation and digital asset manipulation.

%% file: Section/section4.8_novel_view.tex
\subsection{Novel View Synthesis} 
\label{sec:novel_view}

\begin{figure}[h]
  \centering
  \includegraphics[width=\linewidth]{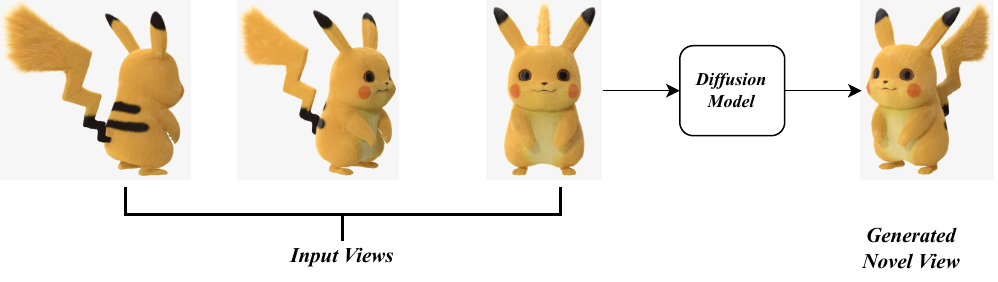}
  \caption{Example of novel view synthesis task.}
  \label{fig:novel_view}
\end{figure}

Novel view synthesis uses diffusion models to generate consistent multi-perspective scene representations, addressing challenges in 3D reconstruction and view extrapolation from limited inputs, as shown in Figure~\ref{fig:novel_view}. 
Recent methods integrate 2D diffusion priors with 3D geometric constraints through various architectural innovations. For example, \citet{nerdi} and \citet{holodiffusion} employ hybrid 2D-3D frameworks where NeRDi optimizes neural radiance fields by minimizing diffusion-based distribution losses across rendered views, while HOLODIFFUSION trains a 3D UNet using explicit-implicit feature grids with 2D supervision to maintain spatial consistency. These methods demonstrate how 3D-aware feature representations can bridge 2D image generation and volumetric scene understanding.

Several studies enhance multi-view consistency via geometric constraints and attention mechanisms. For example, \citet{consistent_view} implement epipolar attention layers guided by camera poses to establish feature correspondences between input and novel views, particularly effective for large camera motions. Similarly, \citet{nvs_diffusion} develop DiM with stochastic conditioning and cross-attention UNet variants to improve 3D consistency during image generation. 
Furthermore, \citet{syncdreamer} achieve synchronized multi-view synthesis through 3D-aware feature attention and joint probability modeling, while \citet{viewdiff} incorporate cross-frame attention and autoregressive strategies for temporal view coherence.

For sparse input scenarios, methodologies combine neural rendering with conditional diffusion. For example, \citet{sparsefusion} utilize view-conditioned latent diffusion models to optimize neural 3D representations from segmented images, whereas \citet{zero1to3} and \citet{zero123++} address single-image reconstruction through synthetic dataset training and multi-view joint distribution modeling. 
The latter introduces scaled reference attention and modified noise schedules to mitigate texture degradation. 
Furthermore, \citet{generative_novel_view} and \citet{nerfdiff} enhance geometric fidelity through 3D feature volumes and NeRF-guided distillation, effectively resolving occlusion ambiguities.

Advanced frameworks leverage cross-modal priors for comprehensive scene generation. And \citet{consistent1to3} propose a hierarchical paradigm combining scene representation transformers with diffusion models for 360° view synthesis, while \citet{viewdiff} integrate text-to-image diffusion models with volumetric rendering for multi-view consistent generation from textual or visual inputs. These approaches demonstrate the potential of combining diffusion-based generation with explicit 3D reasoning to achieve photorealistic and geometrically plausible novel view synthesis across diverse application scenarios.

%% file: Section/section4.9_depth.tex
\subsection{Depth Estimation} 
\label{sec:depth_estimation}

\begin{figure}[h]
  \centering
  \includegraphics[width=\linewidth]{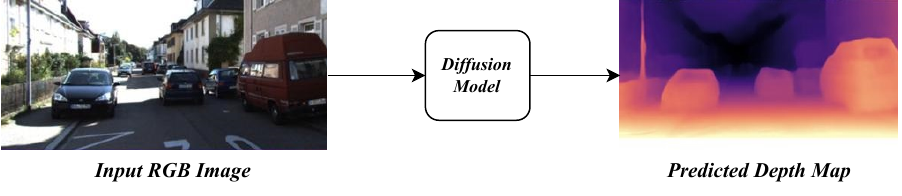}
  \caption{Example of depth estimation task.}
  \label{fig:depth_estimate}
\end{figure}

Depth estimation involves predicting 3-dimensional scene geometry from 2-dimensional visual inputs, such as images or video frames, as shown in Figure~\ref{fig:novel_view}. 
Recent advances employ diffusion models to iteratively enhance depth map quality, producing precise spatial representations vital for applications, including 3D reconstruction, augmented reality, and autonomous navigation systems requiring environmental perception capabilities.

Current methodologies address data scarcity through synthetic data generation using diffusion techniques. For example, \citet{atlantis} developed Depth-2-Underwater ControlNet, which transforms terrestrial depth maps into realistic underwater scenes using stable diffusion while preserving geometric accuracy. And \citet{diffusion_challenge} leverages text-to-image diffusion models with depth-aware control to generate challenging synthetic environments with the corresponding depth annotations. 
And \citet{repurposing} proposes Marigold, a stable diffusion-based framework achieving state-of-the-art monocular depth estimation through synthetic data fine-tuning while retaining pre-trained visual priors by modifying only the denoising U-Net component.

Pre-training methods effectively mitigate data limitations in diffusion-based methods. For example, \citet{ecodepth} incorporates global contextual features of pre-trained Vision Transformers \citep{vit} to improve depth prediction accuracy. And \citet{surprising} introduces DDVM, which combines self-supervised pre-training in image translation tasks with subsequent supervised RGB-D fine-tuning. On the other hand, domain adaptation methods like \citet{sediff}'s framework employ depth-consistent style transfer to extract domain-invariant features, demonstrating improved cross-environment generalization.

Architectural innovations further advance diffusion applications in depth estimation. For example, \citet{ddp_dense} proposes a multitask conditional diffusion framework that achieves competitive benchmark performance without task-specific modifications. And \citet{diff_aug_depth} develops DADP, integrating noise and depth predictors to enhance structural consistency in sparsely annotated autonomous driving scenarios. Furthermore, \citet{diffusiondepth} presents a self-refinement approach in which models learn to reverse diffusion processes from refined depth predictions to random distributions, effectively addressing sparse ground-truth limitations. These methodological developments collectively expand the applicability of diffusion models to diverse depth estimation challenges while reducing the reliance on extensive real-world training data.

%% file: Section/section5_datasets.tex
\section{3D Datasets and Metrics}
\label{sec:dataset}

\subsection{3D Datasets}

\begin{table*}
    \caption{Commonly used 3D datasets.}
    \label{tab:dataset}
    \centering
    \setlength{\tabcolsep}{6.5mm}{\begin{tabular}{lcccc}
        \toprule
        \textbf{Name} & \textbf{Type} & \textbf{Samples}  & \textbf{Category} & \textbf{Source} \\ 
        \midrule
        \midrule
        ShapeNet \citep{shapenet}        & Object  & 51K  & Mesh  & Synth   \\ 
        PartNet \citep{partnet}          & Object  & 574K & Mesh  & Synth  \\ 
        ModelNet40 \citep{modelnet}      & Object  & 12K  & Mesh  & Synth  \\ 
        3dscan \citep{redwood}           & Object  & 10K  & RGBD,PC & Real \\  
        MVP \citep{mvp}                  & Object  & 100K & PointCloud & Synth \\
        Completion3D \citep{completion3d}& Object  & 30K  & PointCloud & Synth \\
        PhotoShape \citep{photoshape}    & Object  & 11K  & Mesh  & Synth \\
        ABO \citep{abo}                  & Object  & 147K & Mesh  & Synth \\
        YCB \citep{ycb}                  & Object  & 600  & RGBD,Mesh & Real \\
        CO3D \citep{co3d}                & Object  & 19K  & MultiView  & Real \\
        Objaverse \citep{objaverse}      & Object  & 800K & Mesh  & Synth \\
        GSO \citep{gso}                  & Object  & 1.0K & Mesh  & Real \\
        Cap3D \citep{cap3d}              & Object  & 785k & Mesh  & Synth \\
        Text2Shape \citep{text2shape}    & Object  & 192K & Mesh  & Synth \\
        3D-FUTURE \citep{3dfuture}       & Object  & 10K & Mesh  & Synth \\
        OmniObject3D \citep{omniobject3d}   & Object & 6K  & Mesh,PointCloud & Real \\ 
        ScanObjectNN \citep{scanobjectnn}   & Object & 2.9K & PointCloud  &  Real \\ 
        \midrule
        Renderpeople \citep{renderpeople}& Human   & 40K & Mesh  & Synth \\
        THuman 2.0 \citep{thuman}        & Human   & 500 & RGBD  & Real \\
        AzurePeople \citep{AzurePeople}  & Human   & 56  & RGBD  & Real  \\
        HumanML3D \citep{humanml3D}      & Human   & 15K & Mesh  & Synth \\
        AMASS \citep{amass}              & Human   & 11K & Mesh  & Synth \\
        HumanAct12 \citep{humanact12}    & Human   & 1.2K & Mesh  & Synth  \\
        UESTC \citep{uestc}              & Human   & 25.6K & RGBD  & Real \\
        FaceScape \citep{facescape}      & Human    & 18.8K & Mesh  & Real \\
        CustomHumans \citep{customhumans}& Human   & 600  & Mesh  & Real \\
        CAPE \citep{cape}                & Human   & 600  & Mesh  & Real \\
        EMDB \citep{emdb}                & Human   & 81   & RGBD    & Real \\
        \midrule
        Replica \citep{replica}          & Scene   & 18   & Mesh  & Real \\
        3D-FRONT \citep{3dfront}         & Scene   & 19K & Mesh  & Synth \\
        KITTI \citep{kitti}              & Scene   & 389  & RGBD,PointCloud & Real \\
        ScanNet \citep{scannet}          & Scene   & 1.5K & RGBD,PointCloud & Real \\
        S3DIS \citep{s3dis}              & Scene   & 271  & PointCloud & Real \\
        AVD \citep{avd}                  & Scene   & 20K  & RGBD,PointCloud  & Real \\
        LLFF \citep{llff}                & Scene   & 32   & MultiView     & Real \\
        DTU \citep{dtu}                  & Scene   & 80   & MultiView     & Real \\
        SceneNet \citep{scenenet}        & Scene   & 57   & RGBD     & Sythn \\
        NYUdepth \citep{nyudepth}        & Scene   & 1.5K & RGBD     & Real  \\
        SUN RGB-D \citep{sunrgbd}        & Scene   & 10K  & RGBD     & Real \\
        nuScenes \citep{nuscenes}        & Scene   & 1K   & PointCloud     & Real \\
        Waymo \citep{waymo}              & Scene   & 2K   & PointCloud     & Real \\
        GL3D \citep{gl3d}                & Scene   & 113  & Mesh     & Real \\
        Realestate10K \citep{Realestate10K} & Scene & 750K & MultiView    & Real \\
        Matterport3D \citep{matterport3D} & Scene  & 90   & RGBD     & Real  \\
        DDAD \citep{ddad}                & Scene   & 254  & RGBD     & Real \\
        Hyper-sim \citep{hypersim}       & Scene   & 461  & Mesh     & Synth \\
        \bottomrule
    \end{tabular}
    }
\end{table*}

\begin{table*}
    \centering
    \caption{Metrics and corressponding descriptions.}
    \label{tab:metrics}
    \begin{tabularx}{\textwidth}{@{}l X l X@{}}
        \toprule
        \multicolumn{4}{c}{\textbf{Distance Metrics}} \\
        \midrule
        \midrule
        CD      & Measures the average difference based on the nearest neighbor distance between point sets.        & TMD & Evaluates the total difference between two distributions in multi-dimensional space. \\
        EMD     & Measures the difference between two mass distributions through the minimum conversion cost.    & LFD     & Quantifies the differences in light fields from different viewpoints. \\
        \midrule
        \multicolumn{4}{c}{\textbf{Coverage Metrics}} \\
        \midrule
        COV     & Measures the proportion of generated samples that cover test samples.            & 1-NNA & Analyzes the mixing degree between test and generated samples based on nearest neighbors. \\
        \midrule
        \multicolumn{4}{c}{\textbf{Distribution Metrics}} \\
        \midrule
        FID     & Compares the similarity of distributions in feature space through mean and covariance. & KID & Measures the distribution distance in feature space based on kernel functions. \\
        FVD     & Extends FID to evaluate the quality of generated video frame features.            & MMD & Measures the distance between distributions by optimal matching of elements. \\
        \midrule
        \multicolumn{4}{c}{\textbf{Similarity Metrics}} \\
        \midrule
        IoU     & Measures the accuracy of detection boxes through overlapping regions.                & PCK  & Evaluates pose estimation accuracy based on the distance between key points and ground truth. \\
        DISTS   & Combines structural and texture similarity to evaluate image quality.          & ReID & Measures the matching capability for person re-identification across images. \\
        \midrule
        \multicolumn{4}{c}{\textbf{Quality Metrics}} \\
        \midrule
        PSNR    & Evaluates signal fidelity based on peak signal-to-noise ratio.              & VQ   & Comprehensive aesthetic and perceptual quality evaluation of generated content. \\
        SSIM    & Measures image similarity through structure, brightness, and contrast.      & SC   & Evaluates the structural integrity of generated content relative to the ground truth. \\
        PF      & Measures the degree of match between generated samples and given prompts.          & PQ   & Evaluates the quality of generated images based on human perception. \\
        \midrule
        \multicolumn{4}{c}{\textbf{Error Metrics}} \\
        \midrule
        REL     & Measures the mean absolute relative error between predicted and ground truth values.            & RMSE     & Measures prediction bias through root mean square error. \\
        SqRel   & Measures the mean squared relative error between predicted and ground truth values.            & RMSElog  & Root mean square error on a logarithmic scale. \\
        \bottomrule
    \end{tabularx}
\end{table*}

In this section, we summarize some of the most commonly used datasets in current diffusion-related 3D tasks. We categorize them into three groups: object-based, human-based, and scene-based. 
All datasets and their details are listed in Table \ref{tab:dataset}.

\subsubsection{Object Datasets}

Among the various datasets available for 3D object dataset, ShapeNet \citep{shapenet} stands out as one of the earliest and most foundational, featuring 51K synthetic 3D CAD models across diverse categories. It has significantly influenced subsequent research in the field. 
CO3D \citep{co3d} is notable for its large scale real scanned 3D objects, comprising 19K models, which allows for detailed analysis and has become a vital resource for fine-grained 3D object-related tasks. 
On the other hand, Objaverse \citep{objaverse} represents a major advancement in dataset size, with nearly 800K models, making it the largest among the listed datasets and facilitating extensive applications in training deep learning models for 3D understanding.
In addition to the above datasets, other datasets such as ABO \citep{abo} and GSO \citep{gso} also play an important role in the research of 3D object generation.

\subsubsection{Human Datasets}

Within those datasets focused on human representation, Renderpeople \citep{renderpeople} is notable for being the largest synthetic dataset, comprising nearly 40K models. Its extensive scale makes it a valuable resource for various applications in computer graphics and machine learning. In contrast, the THuman 2.0 dataset \citep{thuman}, with 500 real human models, is recognized as one of the earliest datasets in this field, contributing significantly to research on human motion and behavior recognition. Another important dataset is UESTC \citep{uestc}, featuring 25.6K real human models, which stands out for its substantial size and practical applications in real-world scenarios. 
Other datasets, such as HumanML3D \citep{humanml3D}, AMASS \citep{amass} and FaceScape \citep{facescape}, also contribute important insights into human modeling and animation.

\subsubsection{Scene Datasets}

In those 3d scene datasets, Realestate10K \citep{Realestate10K} is notable for being the largest dataset, comprising 750K scenes, which significantly contributes to advancements in scene recognition and analysis. 
KITTI \citep{kitti} holds the distinction of being one of the earliest datasets established for autonomous driving applications, encompassing 389 scenes and providing essential data for research in computer vision and robotics. 
ScanNet \citep{scannet} is recognized for its extensive real-world indoor scene collection, featuring nearly 1,500 diverse scenes, making it one of the most widely utilized datasets in training and benchmarking algorithms. 
Other datasets, such as NyuDepth \citep{nyudepth} and Waymo \citep{waymo}, also play important roles in 3D indoor and outdoor scene-related researches.

\subsection{Evaluation Metrics}




We categorise and list all the commonly used metrics in 3D-related tasks in Table \ref{tab:metrics}, including Chamfer Distance (CD), Earth Mover’s Distance (EMD), Total Mutual Difference (TMD), 
Light Field Distance (LFD), Coverage (COV), Nearest Neighbour (1-NNA), 
Frechet Inception Distance (FID), Frechet Video Distance (FVD), Kernel Inception Distance (KID), Minimum Matching Distance (MMD),
Intersection over Union (IoU), Deep Image Structure and Texture Similarity (DISTS), Percentage of Correct Keypoints (PCK), Reidentification Score (ReID), 
Peak Signal-to-Noise Ratio (PSNR), Structural Similarity Index (SSIM), Prompt Fidelity (PF), Visual Quality (VQ), Structure Completeness (SC), Perceptual Quality (PQ), 
Mean Absolute Relative Error (REL), Mean Squared Relative Error (SqRel), Root Mean Squared Error (RMSE), Root Mean Squared Log Error (RMSElog). 

%% file: Section/section6_future.tex
\section{Limitations and Future Directions}
\label{sec:future}

\paragraph{Computational Efficiency and Inference Speed} 
Although diffusion models have shown great promise in generating high-quality 3D data, their computational demands remain a significant challenge. These models typically require a large number of iterations to produce results, which increases both the training time and the inference speed. This problem becomes more acute when working with high-dimensional 3D data, where memory and computational requirements scale rapidly. 
Future research could focus on optimizing the number of diffusion steps or exploring more efficient architectures to accelerate inference without sacrificing output quality.

\paragraph{Multimodal Fusion} 
A key challenge in 3D vision tasks is effectively integrating different data modalities, such as 2D images, 3D geometry, and textual descriptions. 
Although diffusion models have been applied successfully to single-modality tasks, there is still room for improvement in multimodal fusion. Future work could explore methods for better combining these modalities within the diffusion framework. For example, employing mechanisms of cross-attention or using embeddings that capture the relationships between 2D, 3D, and textual input could enhance the model’s ability to synthesize richer, more context-aware 3D content. Additionally, developing unified architectures capable of jointly handling multiple modalities could broaden the application of diffusion models in tasks such as 3D scene generation.

\paragraph{Large-scale Pretraining and Transfer Learning} 
Large-scale pre-training has been pivotal in achieving state-of-the-art results in many 2D vision tasks, yet its application in 3D diffusion models remains underexplored, due to the scarcity of 3D datasets and high computational demands. 
Given this situation, effective transfer learning techniques could be developed to allow 2D pre-trained models to generalize across diverse 3D tasks, such as reconstruction, generation, and recognition. Identifying robust pre-training and transfer learning strategies and understanding the transferability of knowledge across 2D and 3D domains are promising areas for future research.

\paragraph{Interpretability and Fine-grained Control} 
One of the major limitations of current diffusion models is the lack of interpretability and fine-grained control over the generation process. 
This issue is particularly relevant in 3D vision, where users may want to influence specific aspects of the generated 3D content, such as shape, texture, or pose. 
Future work should focus on improving the transparency of these models by developing interpretable latent spaces or integrating control mechanisms. Techniques such as latent space manipulation, conditional generation, or disentanglement of various attributes could empower users to direct the output in more predictable ways. 
Moreover, explainability tools could help users understand how the model arrives at specific outputs, increasing trust and usability in real-world applications.

\paragraph{Complex and Dynamic Scenes} 
Current diffusion models in 3D vision are mostly applied to static objects or scenes, but extending them to handle complex and dynamic environments remains a key challenge. 
Real-world 3D scenes often involve dynamic elements such as moving objects, changing lighting conditions, or large-scale outdoor environments. To address these complexities, future research should explore methods for encoding temporal information and handling dynamic changes within diffusion models. 
This could involve the integration of spatiotemporal diffusion processes or the development of hybrid models that combine static scene understanding with temporal dynamics. Successfully extending these models to dynamic and large-scale 3D environments would open up new possibilities for applications in simulation, virtual reality, and robotics.

\paragraph{Physical Constraints} 
Adding physical constraints to 3D diffusion models is crucial for generating realistic and plausible 3D content. Future research could focus on integrating principles from physics, such as conservation laws, collision detection, and material properties, to ensure that generated 3D objects behave in accordance with the real world.
This could involve embedding physics-based priors or loss functions into the training process to encourage physically accurate outputs. For example, enforcing geometric consistency or simulating real-world interactions, such as gravity or fluid dynamics, could enhance the realism of the generated 3D content. 
By grounding these models in physical laws, they could be applied more effectively in engineering, simulation, and other fields where physical accuracy is paramount.





%% file: Section/section7_conclusion.tex
\section{Conclusion}
\label{sec:conclusion}

In this paper, we present a comprehensive survey on the application of diffusion models to 3D vision tasks. We explore how these models, originally designed for 2D generative tasks, have been adapted to handle the complexities of 3D data, such as point clouds, meshes, and voxel grids. 
We review key approaches that apply diffusion models to tasks like 3D object generation, shape completion, point cloud reconstruction, and scene generation.
Moreover, we provide an in-depth discussion of the underlying mathematical principles of diffusion models, outlining their forward and reverse processes, as well as the various architectural advancements that enable these models to work with 3D data. 
Additionally, we categorize and analyze the different 3D tasks where diffusion models have shown significant impact, such as text-to-3D generation, mesh generation, and novel view synthesis.
The paper also address the key challenges in applying diffusion models to 3D vision, such as handling occlusions, varying point densities, and the computational demands of high-dimensional data. 
To guide future research, we discuss potential solutions, including improving computational efficiency, enhancing multimodal fusion, and exploring the use of large-scale pretraining for better generalization across 3D tasks.
By consolidating the current state of diffusion models in 3D vision and identifying gaps and opportunities, this paper serves as a foundation for future exploration and development in this rapidly evolving field.